\documentclass[10pt,twocolumn,twoside]{IEEEtran}
\usepackage{latexsym}
\usepackage{amsmath}
\usepackage{amsfonts}
\usepackage{amssymb}
\usepackage{graphicx}

\newtheorem{theorem}{Theorem}

\newtheorem{corollary}{Corollary}
\newtheorem{proposition}[theorem]{Proposition}
\newcommand{\commentout}[1]{}

\title{Anomaly Detection and Removal Using Non-Stationary Gaussian Processes}

\author{Steven Reece, Roman Garnett, Michael Osborne and Stephen Roberts\\ Robotics Research Group\\
  Dept. Engineering Science\\ Oxford University, UK.}

\date{}

\begin{document}

\maketitle
\pagestyle{empty}

\section*{abstract}
This paper proposes a novel Gaussian process approach to fault removal
in time-series data.  {\it Fault removal} does not delete the faulty
signal data but, instead, massages the fault from the data.  We assume
that only one fault occurs at any one time and model the signal by two
separate non-parametric Gaussian process models for both the physical
phenomenon and the fault.  In order to facilitate fault removal we
introduce the {\it Markov Region Link} kernel for handling
non-stationary Gaussian processes.  This kernel is piece-wise
stationary but guarantees that functions generated by it and their
derivatives (when required) are everywhere continuous.  We apply this
kernel to the removal of drift and bias errors in faulty sensor data
and also to the recovery of EOG artifact corrupted EEG signals.

\section{Introduction}
\label{sec:gaussianprocesses}
Gaussian processes (GPs) are experiencing a resurgence of interest.
Current applications are in diverse fields such as geophysics, medical
imaging, multi-sensor fusion~\cite{osborne08} and sensor
placement~\cite{garnett10}.  A GP is often thought of as a ``Gaussian
over functions''~\cite{rasmussen06}.  It can be used to construct a
distribution over functions via a prior on the function's values.
The prior is specified through a positive-definite kernel, which
determines the covariance between two outputs as a function of their
corresponding inputs. A GP is fully described by its mean and
covariance functions.  Suppose we have a set of training data
$D=\{(x_1,y_1),\ldots,(x_n,y_n)\}$ drawn from a noisy process:
\begin{align}
y_i=f(x_i)+\epsilon_i
\label{basicf}
\end{align}
where $f(x_i)$ is the {\it real process} and $\epsilon_i$ is zero-mean
Gaussian with variance $\sigma^2$.  For convenience both inputs and
outputs are aggregated into $X=\{x_1,\ldots,x_n\}$ and
$Y=\{y_1,\ldots,y_n\}$ respectively.  The GP estimates the value of
the function $f$ at sample locations $X_*=\{x_{*1},\ldots,x_{*m}\}$.
The basic GP regression equations are given in \cite{rasmussen06}:
\begin{small}
\begin{align}
\hat{f}_*=&K(X_*,X)[K(X,X)+\sigma^2_nI]^{-1}Y\ ,\label{GPpred1} \\
{\rm Cov}(f_*)=& K(X_*,X_*)-K(X_*,X)[K(X,X)+\sigma^2_nI]^{-1}K(X_*,X)^T\label{GPpred2}
\end{align}
\end{small}
where $\hat{f}_*$ is the marginal posterior mean at $X_*$ and
${\rm Cov}(f_*)$ is the corresponding covariance.  The {\it prior}
covariance at $X_*$ is $K(X_*,X_*)$ where the covariance matrix
$K(X_*,X_*)$ has elements $K_{ij}=\Bbb{K}(x_i,x_j;\theta_k)$.  The
function $\Bbb{K}$ is called the {\it kernel function}.  These
functions come with parameters, $\theta_k$, which can be tuned to suit
the application.  The kernel function is chosen according to known
properties of the underlying processes, such smoothness and
stationarity.  There are many existing kernels to choose from (see,
for example,~\cite{rasmussen06}).  We will develop a new kernel for
our fault recovery algorithm.  The prior mean is traditionally set to
zero and we follow this convention.  However, the results in this
paper can be readily generalised to non-zero prior means.  The term
$\sigma_n^2 I$ captures the noise in Eqn~\ref{basicf}.

The GP parameters $\theta$ (which includes $\sigma$ and
hyperparameters associated with the covariance function) can be
inferred from the data through Bayes' rule
\begin{eqnarray} 
p(\theta\mid Y,X)=\frac{p(Y\mid X,\theta)}{p(Y\mid X)}p(\theta)\ .
\label{learning}
\end{eqnarray} 
The parameters are usually given a vague prior distribution
$p(\theta)$. 

The paper is organised as follows.  Section~\ref{sec:fault_recovery}
demonstrates how fault recovery can be achieved using non-stationary
Gaussian processes.  Section~\ref{sec:non-stat} reviews current
covariance functions for modelling non-stationary processes.  Then,
Section~\ref{sec:probdesc} presents the problem description as a
piece-wise stationary problem with boundary constraints at the onset
of faults.  Section~\ref{sec:MRL} presents the MRL kernel for
functions which are continuous boundaries and this is extended to
cases where function derivatives are continuous at region boundaries
in Section~\ref{sec:derivbnd}.  In Sections~\ref{sec:appl1}
and~\ref{sec:appl2} we demonstrate the efficacy of our approach on
simulated data from a faulty sensor target estimation problem as well
as a dataset involving EOG artifact corrupted EEG signals.  Finally,
we conclude in Section~\ref{sec:conclusions}.

\section{Gaussian Process Faulty Recovery}
\label{sec:fault_recovery}
We assume that a faulty signal is the linear composition of the real
physical phenomenon of interest and a deterministic offset induced by
a fault either in the sensor measurement or in the physical phenomenon
itself.  Again, the signal is assumed to be drawn from a noisy
process:
\begin{align}
y_i=f(x_i)+e(x_i)+\epsilon_i
\label{basicf}
\end{align}
where $f(x_i)$ is the {\it real process}, $e(x_i)$ is the {\it fault
  process} and $\epsilon_i$ is zero-mean Gaussian with variance
$\sigma^2$.

Again, both inputs and outputs are aggregated into
$X=\{x_1,\ldots,x_n\}$ and $Y=\{y_1,\ldots,y_n\}$ respectively.  The
GP regression equations, (\ref{GPpred1}) and (\ref{GPpred2}), are
modified so that both the real and fault processes can be inferred
separately:
\begin{small}
\begin{align*}
\hat{f}_*=&K_f(X_*,X)[K_s(X,X)+\sigma^2_nI]^{-1}Y\ ,\\
{\rm Cov}(f_*)=& K_f(X_*,X_*)-K_f(X_*,X)\times\\
&\hspace*{2cm}[K_s(X,X)+\sigma^2_nI]^{-1}K_f(X_*,X)^T\ ,\\ \ \\
\hat{e}_*=&K_e(X_*,X)[K_s(X,X)+\sigma^2_nI]^{-1}Y\ ,\\
{\rm Cov}(f_*)=& K_e(X_*,X_*)-K_e(X_*,X)\times\\
&\hspace*{2cm}[K_s(X,X)+\sigma^2_nI]^{-1}K_e(X_*,X)^T
\end{align*}
\end{small}
where $K_s(X,X)=K_f(X,X)+K_e(X,X)$.  $\hat{f}_*$ is the marginal
posterior mean at $X_*$ and ${\rm Cov}(f_*)$ is the corresponding
covariance of the real process.  $\hat{e}_*$ and ${\rm Cov}(e_*)$ are
the marginal posterior mean and corresponding covariance of the fault
process, respectively.  The {\it prior} covariances at $X_*$ are
$K_f(X_*,X_*)$ and $K_e(X_*,X_*)$ for the real process and the fault
process, respectively, where, again, each covariance matrix is
generated by a kernel function.  We may assign different kernels to
the real and fault processes.

\begin{figure}[ht]
\begin{center}
\begin{tabular}{cc}
\includegraphics[width=0.23\textwidth]{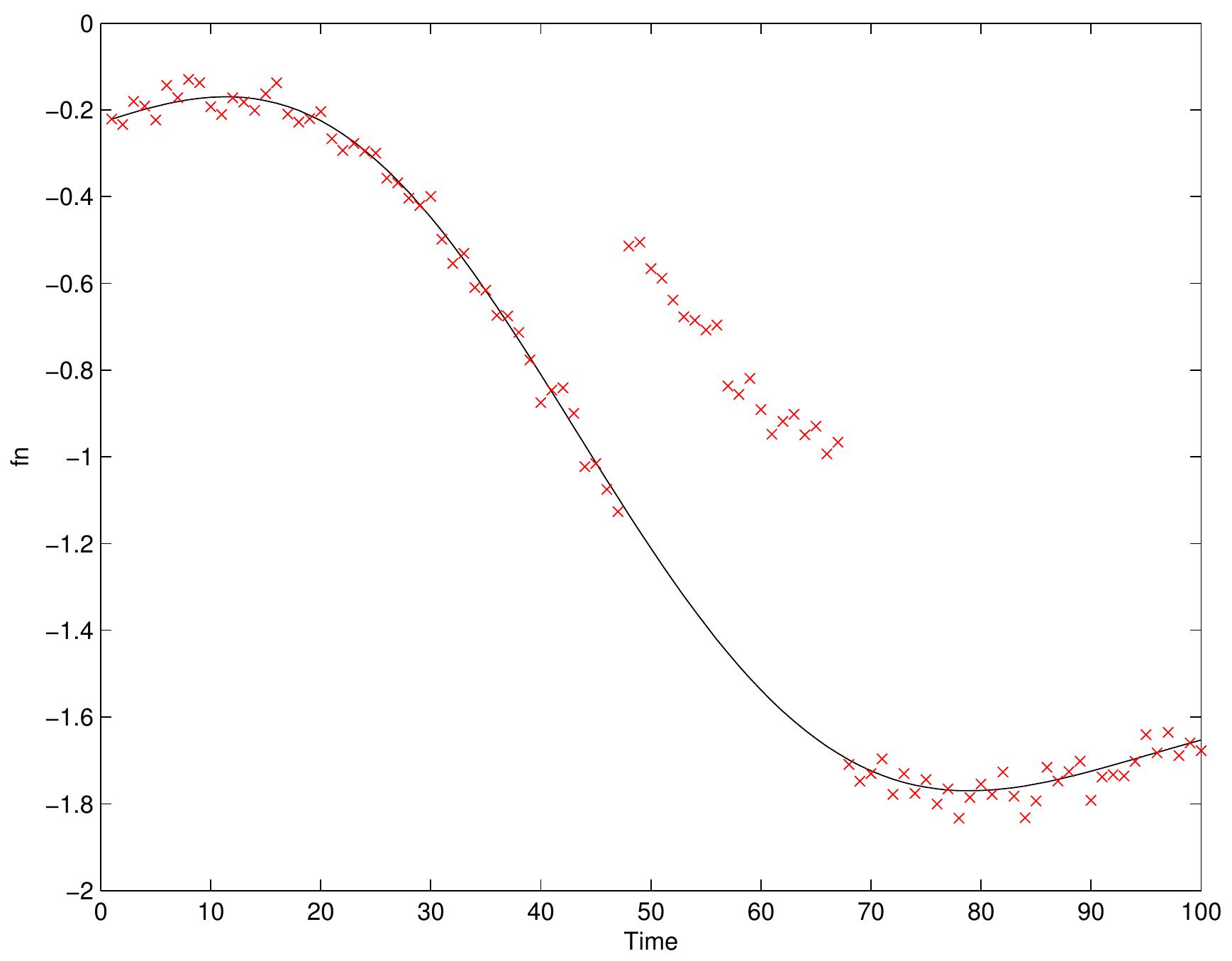}&
\includegraphics[width=0.23\textwidth]{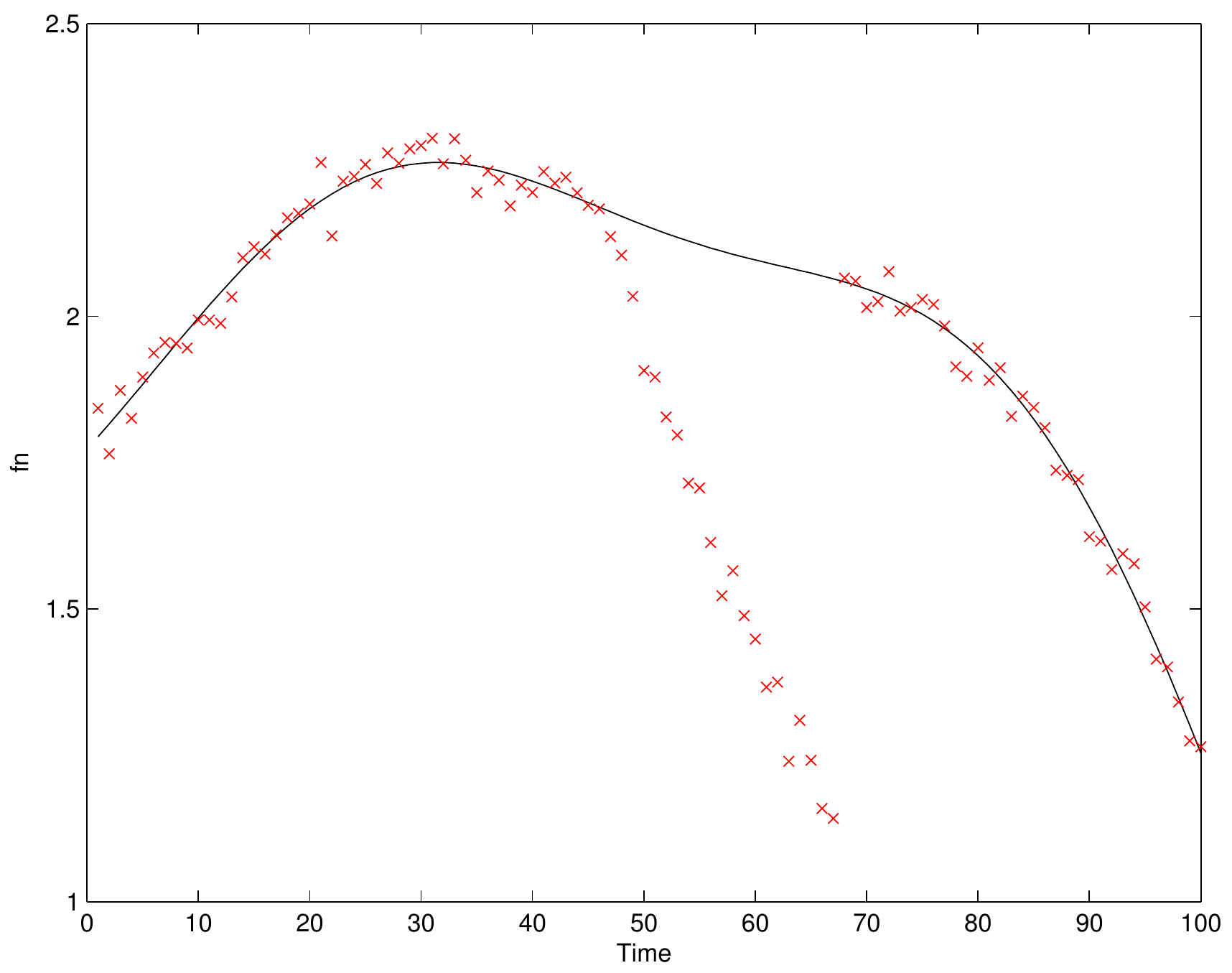}\\
 (a) Bias & (b) Drift  
\end{tabular}
\caption{\label{fault1} Typical real process and faulty observations  
for both bias and drift faults.}
\end{center}
\end{figure}

We shall consider two kinds of faults in this paper, {\it drift} and
{\it bias} (see Figure~\ref{fault1}).  Although, our approach can be
extended to arbitrary fault types.  Both faults are temporary, so they
have a start time and an end time.  The fault process is assumed to be
zero outside of this time interval.  Drift faults are gradual.
Starting from zero error they grow over time and then either shrink
back to zero gradually or snap back to zero instantaneously.  Bias
faults are severe and immediate.  They induce a significant error in
the signal at onset which persists until the fault subsides at which
point the signal snaps back onto the real process. The drift fault is
continuous at the start whereas the bias fault is discontinuous.
 
In order to model the drift fault we develop a kernel which guarantees
that the fault is continuous at onset.  We call this kernel the {\it
  continuous, conditionally independent} (or CCI) kernel.  The CCI
kernel has applications outside of fault recovery.  This
non-stationary kernel splices two or more locally stationary kernels
together whilst preserving function continuity throughout.
Consequently, we present this kernel thoroughly for the first time.

\section{Non-Stationary Kernels}
\label{sec:non-stat}
Many applications use stationary covariance functions for which the
kernel is a function of the distance between the input points.
Stationary covariance functions are appealing due to their intuitive
interpretation and their relative ease of construction.
Unfortunately, stationary GP functions are not applicable in
applications where there are input-dependent variations in the model
hyperparameters (e.g. length-scale, amplitude) and kernel families.
Consequently, non-stationary GP functions have been proposed, such as
the neural network kernel~\cite{williams98} and the Gibbs
kernel~\cite{gibbs97}.  

Methods for deriving non-stationary kernels from stationary kernels
have also been proposed.  Perhaps the earliest approach was to assume
a stationary random field within a moving window~\cite{haas90}.  This
approaches works well when the non-stationarity is smooth and gradual.
It fails when sharp changes in the kernel structure occur.  An
alternative solution is to introduce an arbitrary non-linear mapping
(or warping) $u(x)$ of the input $x$ and then apply a stationary
covariance function in the $u$-space~\cite{sampson92}.  Unfortunately,
this approach does not handle sharp changes between different locally
applied kernels very well~\cite{kim05}.  The {\it mixture of GPs}
\cite{tresp00} approach uses the EM algorithm to simultaneously assign
GP mixtures to locations and optimise their hyperparameters.  Although
the mixture approach can use arbitrary local GP kernels, it does not
guarantee function continuity over GP kernel transitions.
Paciorek~\cite{paciorek04} proposes a non-stationary GP kernel which
guarantees continuity over region boundaries.  Unfortunately, this
approach requires that the local, stationary kernels belong to the
same family.

\subsection{Example: The Gibbs Non-Stationary Kernel}
\label{gibbssec}
\begin{figure}[ht]
\begin{center}
\begin{tabular}{cc}
\includegraphics[width=0.23\textwidth]{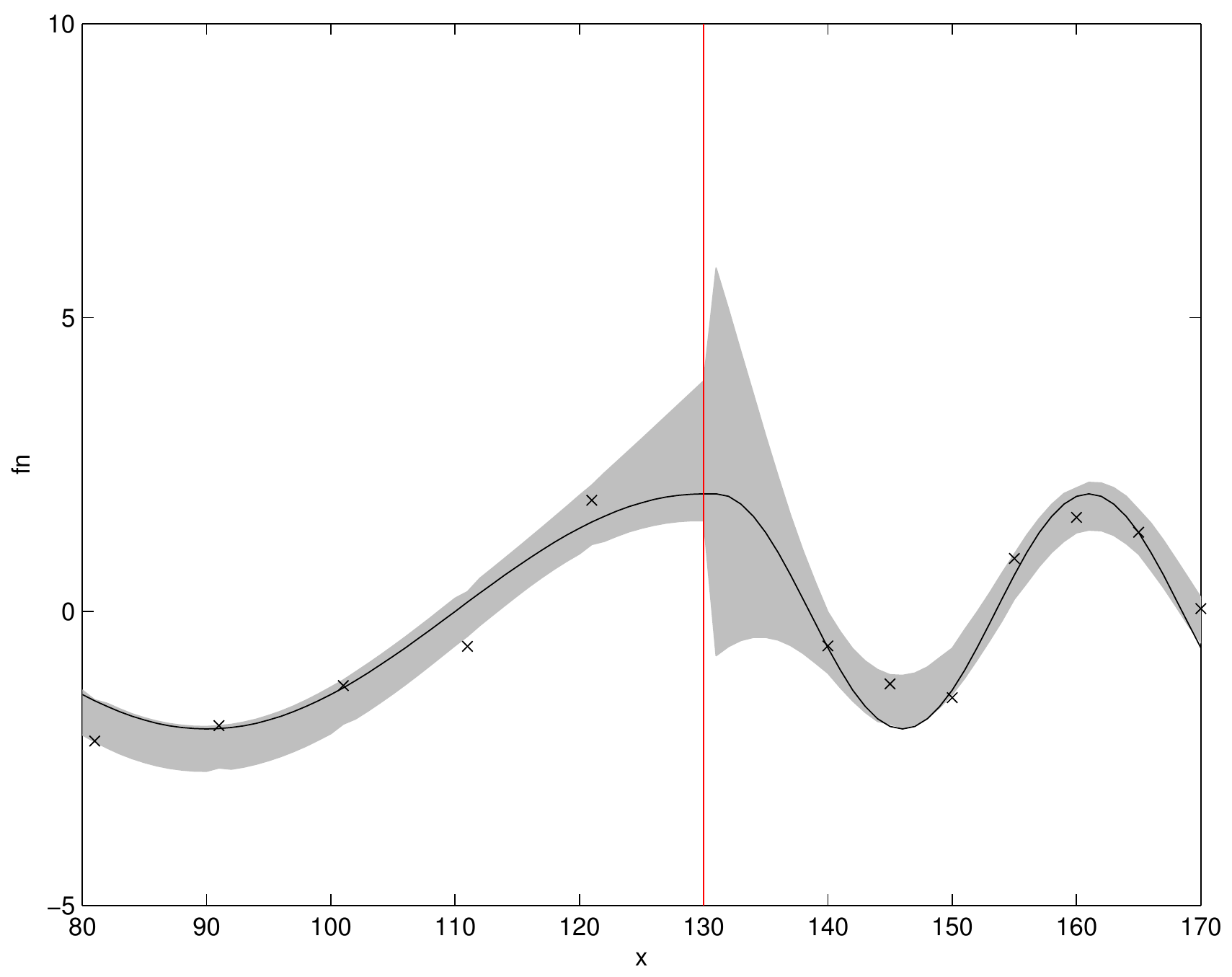}&
\includegraphics[width=0.23\textwidth]{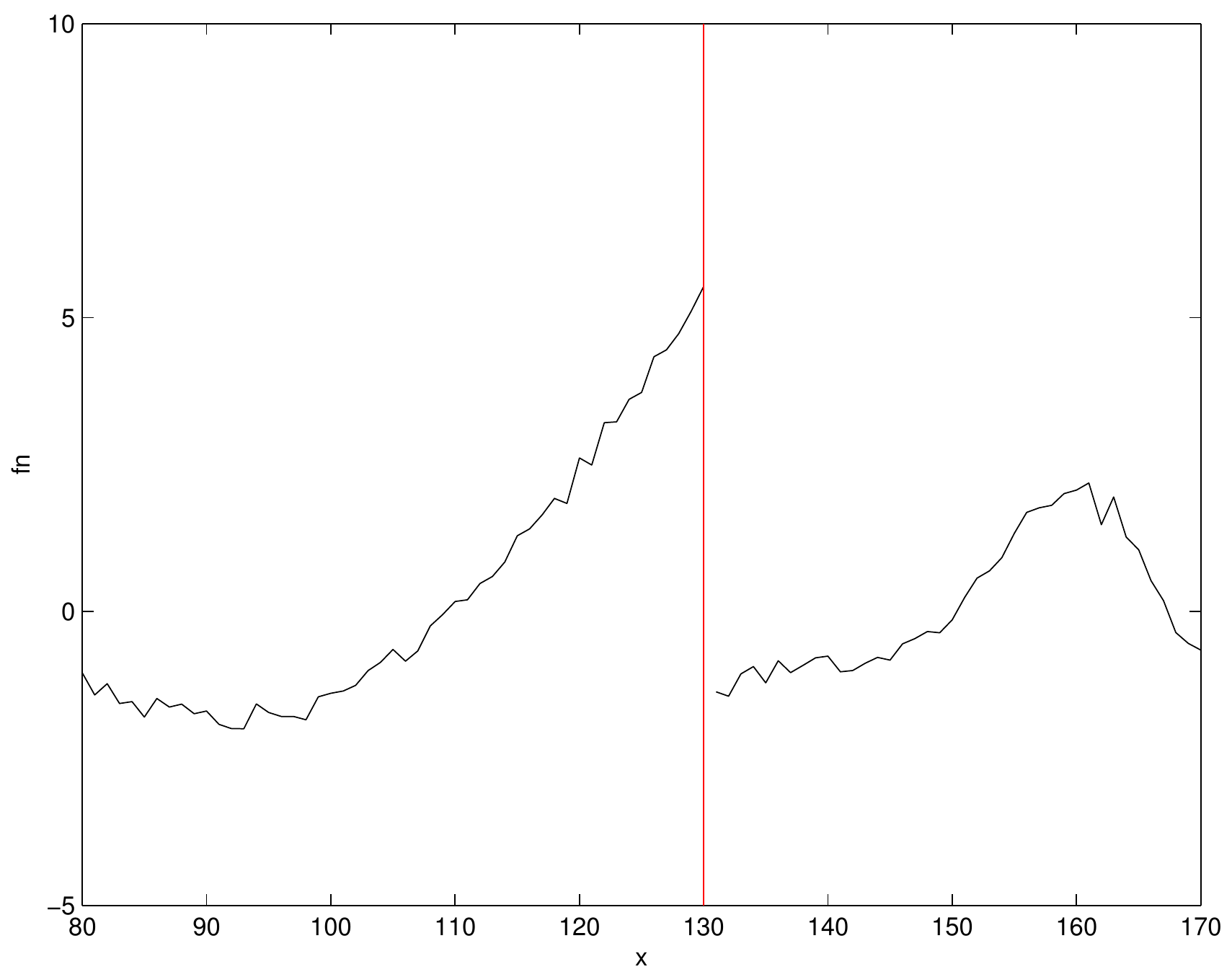}\\
(a) & (b)
\end{tabular}
\end{center}
\caption{\label{exgibbs1} Part (a) shows a function and its estimate
  obtained using the Gibbs Non-stationary Gaussian process
  kernel. Also, (b) a random sample drawn from the Gibbs
  non-stationary kernel typically showing a discontinuity where the
  length-scale changes.}
\end{figure}

Gibbs~\cite{gibbs97,rasmussen06} derived the covariance function:
\begin{small}
\begin{align}
K(x,x^\prime)=\prod_{d=1}^D\left(\frac{2l_d(x)l_d(x^\prime)}{l^2_d(x)+l^2_d(x^\prime)}\right)^{1/2}\exp\left(-\sum_{d=1}^D\frac{(x_d-x^\prime_d)^2}{l^2_d(x)+l^2_d(x^\prime)}\right)
\label{gibbs}
\end{align}
\end{small}
where each length-scale, $l_i(x)$, is an arbitrary positive function of
$x$ and $D$ is the dimensionality of $x$.  If the length-scale varies
rapidly then the covariance drops off quite sharply due to the
pre-factor in Eqn~\ref{gibbs}.  As a consequence the inferred function
estimate can be quite uncertain at length-scale boundaries.  This is
demonstrated in Figure~\ref{exgibbs1}(a) for which the length-scale
changes from $l(x)=35$ for $x\le 130$ to $l(x)=15$ for $x>130$.
Further, the Gibbs kernel does not guarantee that functions generated
by the kernel are continuous.  Figure~\ref{exgibbs1}(b) shows a typical
sample drawn from the posterior Gaussian distribution represented in
Figure~\ref{exgibbs1}(a).

\subsection{Example: Warping of the Input Space}
\label{warpingsec}
\begin{figure}[ht]
\begin{center}
\includegraphics[width=0.45\textwidth]{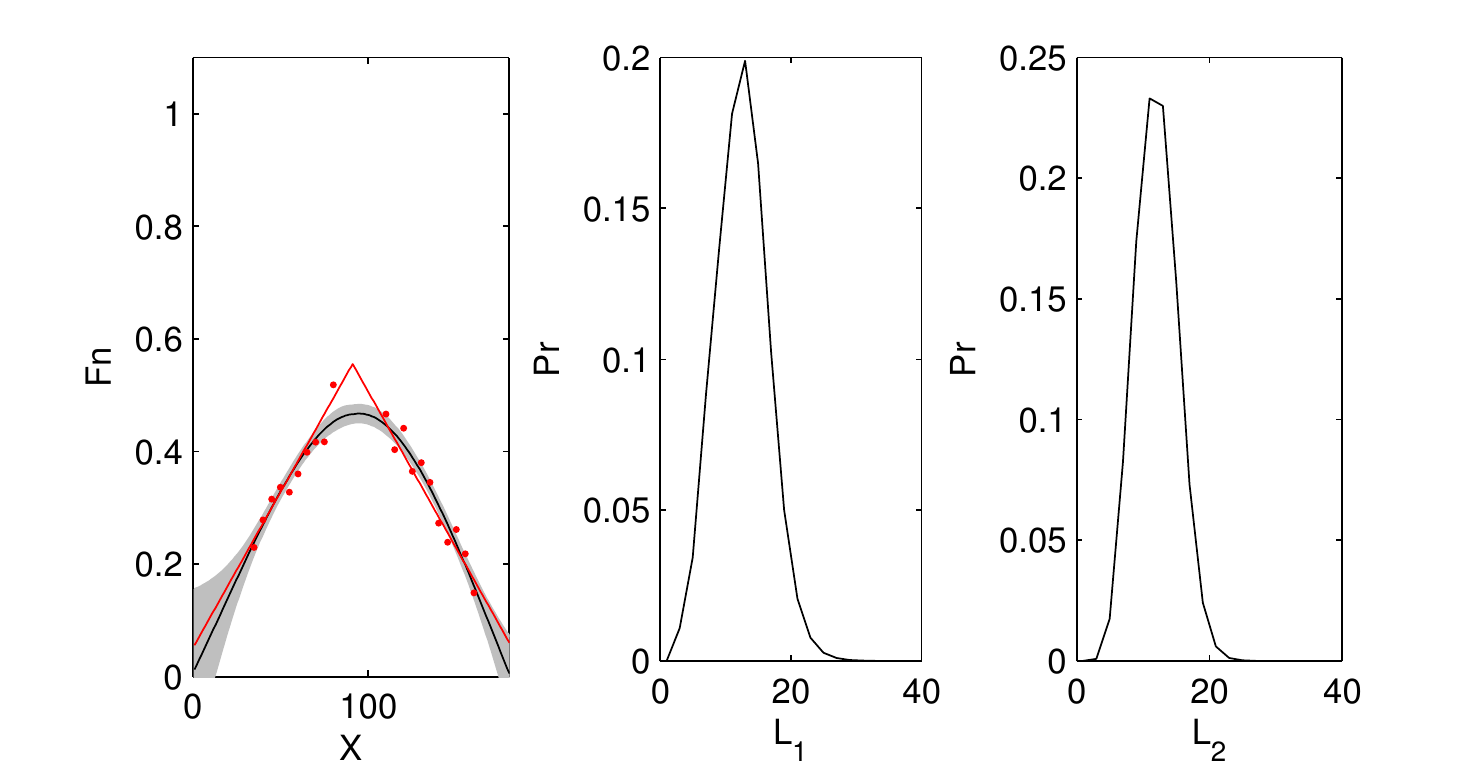}
\includegraphics[width=0.45\textwidth]{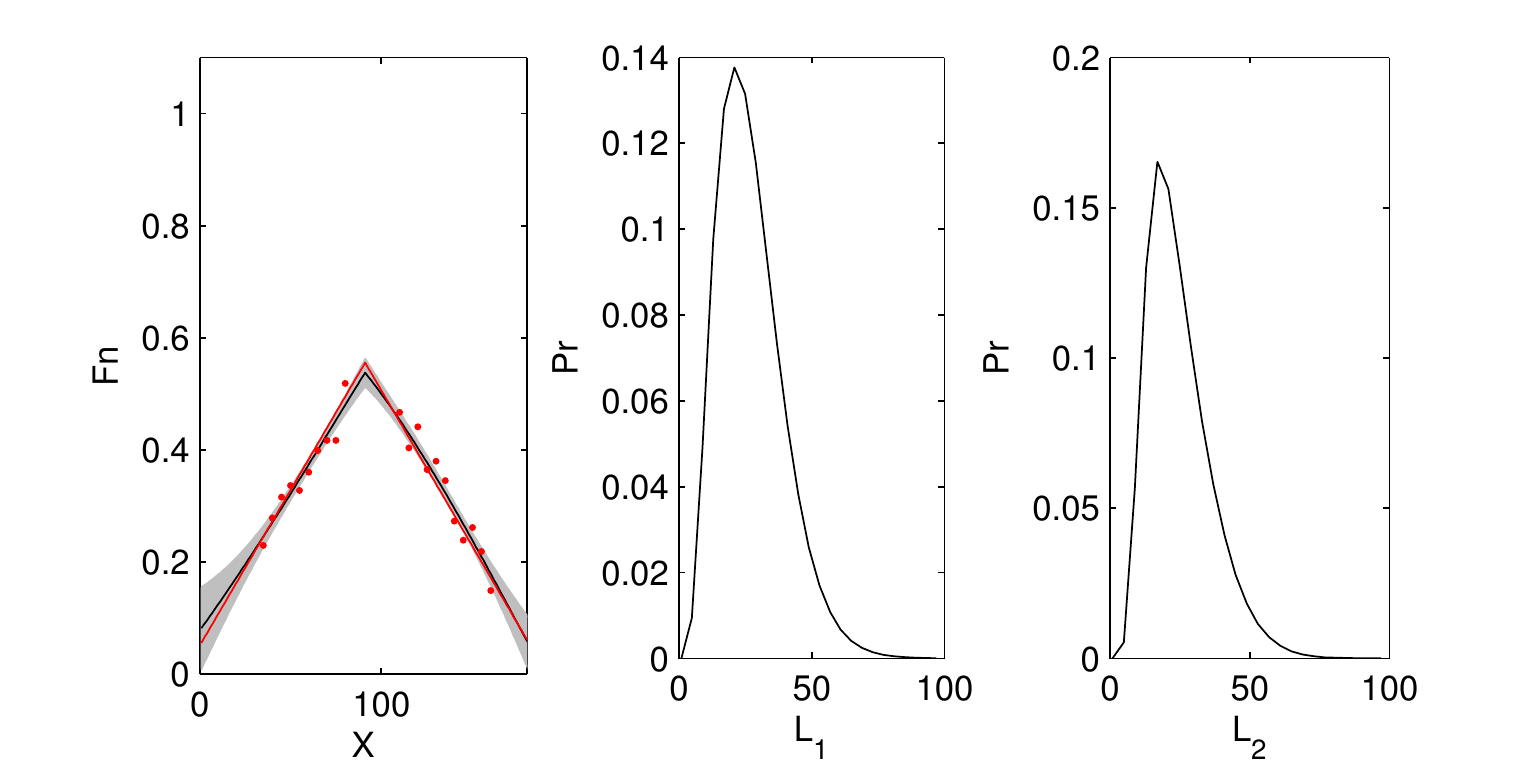}\\
\end{center}
\caption{\label{exwarping2} Low SNR: The left panel shows the ground truth (red
  line), observations (red dots) and GP estimate with mean (black
  line) and first standard deviation (grey region). The other panels show the
distributions over length scales, $L_1$ and $L_2$, inferred for the piece-wise plots each side of $X=100$.}
\end{figure}

\begin{figure}[ht]
\begin{center}
\includegraphics[width=0.45\textwidth]{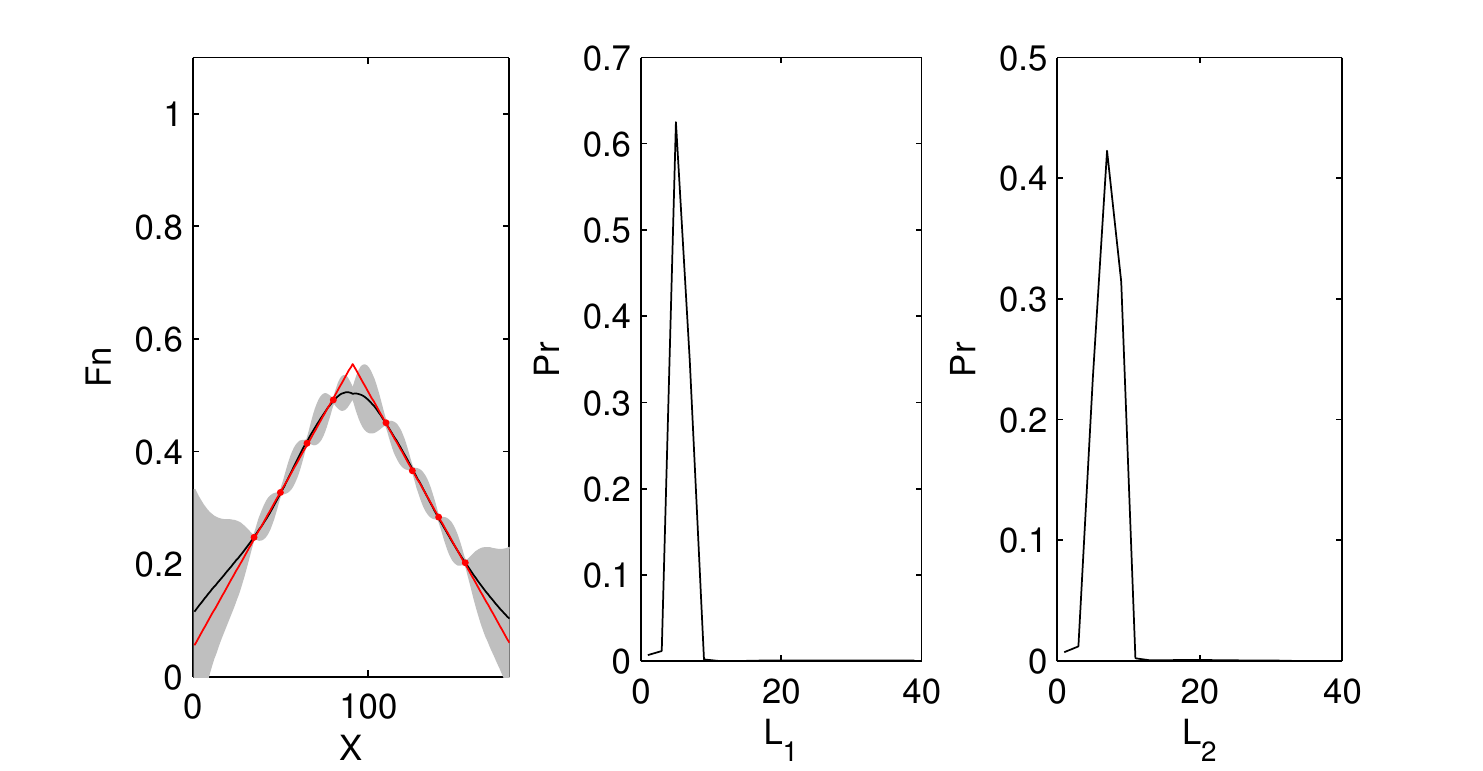}
\includegraphics[width=0.45\textwidth]{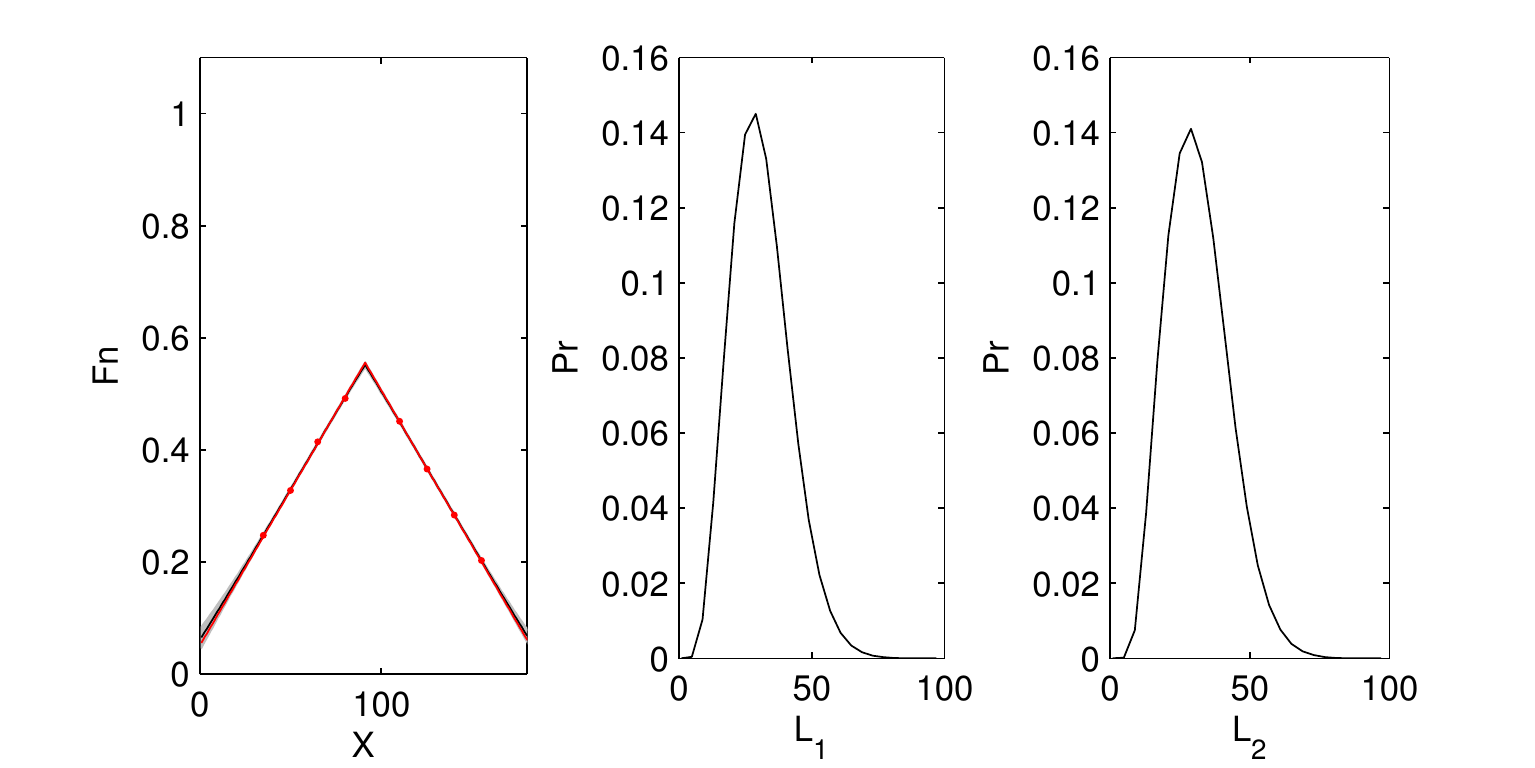}\\
\end{center}
\caption{\label{exwarping3} High SNR: The left panel shows the ground
  truth (red line), observations (red dots) and GP estimate with mean
  (black line) and first standard deviation (grey region). The other
  panels show the distributions over length scales, $L_1$ and $L_2$,
  inferred for the piece-wise plots each side of $X=100$.}
\end{figure}

This example demonstrates the limitation of modelling piece-wise
stationary functions by warping the input space as proposed
by~\cite{sampson92}.  Figures~\ref{exwarping2} and~\ref{exwarping3}
show a continuous wedge function with low and high signal-to-noise
(SNR) respectively.  The figures also show the mean and first standard
deviation of two models:
\begin{itemize}
\item a warped squared exponential
\item two functions, each generated from a squared exponential kernel
  each side of a boundary at $x=100$, and continuous there.
\end{itemize}

For low SNR the warping approach can smooth over features of high
curvature such as the wedge apex.  For high SNR the warping kernel
produces a ``bell'' estimate as it is forced to fit a smooth kernel at
the apex.

In many applications a completely different GP function may be
required to model different regions within the space of interest and
the family of non-stationary covariance functions currently available
in the literature may be too restrictive to model these problems
especially when there are function continuity conditions at region
boundaries.  We will show how arbitrary stationary GP kernels can be
combined to form non-stationary GP covariance priors which preserve
function continuity.  We shall call the new kernel the {\it Markov
  Region Link} (MRL).

\section{Problem Description}
\label{sec:probdesc}
We will assume that a domain can be partitioned such that within
regions (tiles) the process is stationary~\cite{kim05}. Furthermore,
each region may be modelled by kernels from different families.  For
example, one region may be modelled by a Mat\'ern kernel whereas a
neighbouring region may be modelled by a mixture of squared
exponential and periodic kernels.  We do not assume that the functions
generated by these kernels are independent between regions and,
although we desire sharply changing GP kernels or hyperparameters at
region boundaries, we would also like to preserve function continuity
at the boundaries.

Two regions are labelled $R_1$ and $R_2$ and collectively they form
the global region $R$.  A function over $R$ is inferred at sample
locations $X_*=\{x_{*1},\ldots,x_{*m}\}$ given training data at
locations $X=\{x_1,\ldots,x_n\}$.  However, the training data
locations are partitioned between the regions and the region
boundary.~\footnote{In 1D problems the region boundary is often
  referred to as the {\it change-point}.}  Let $X_r$ be the locations
internal to region $R_r$ and let $X_B$ be the boundary.  Then:
\begin{align*}
X=X_1\cup X_B\cup X_2\ .
\end{align*}
We will assume that the function can be modelled using a {\it
  stationary} GP in each region and endeavour to design a global GP
covariance prior which preserves the individual region kernels.  We
will also endeavour to preserve function continuity where desired
across region boundaries including, for example, function continuity
or continuity of function derivatives.  Thus, the necessary conditions
are:
\begin{enumerate}
\item The global kernel $K$ preserves the individual region
  kernels, $K_r$.  That is, $K(X,X)=K_r(X,X)$ for all $X\subseteq
  X_r\cup X_B$ and all regions $r$.
\item The global kernel preserves function continuity, or derivative
  continuity, at the boundary.
\end{enumerate}

\begin{proposition}
\label{prop1}
If two regions, labelled $1$ and $2$, are joined at the boundary $X_B$
and a function defined over the regions is modelled by $K_1$ in region
$R_1$ and $K_2$ in $R_2$ and the function is continuous at the boundary
then:
\begin{align*}
K_1(X_B,X_B)=K_2(X_B,X_B)\triangleq K_B\ .
\end{align*}
\end{proposition}
The boundary covariance, $K_B$, is a hyperparameter which can be
learned from the training data.

\section{The Markov Link Kernel}
\label{sec:MRL}
\begin{figure}[ht]
\begin{center}
\includegraphics[width=0.3\textwidth]{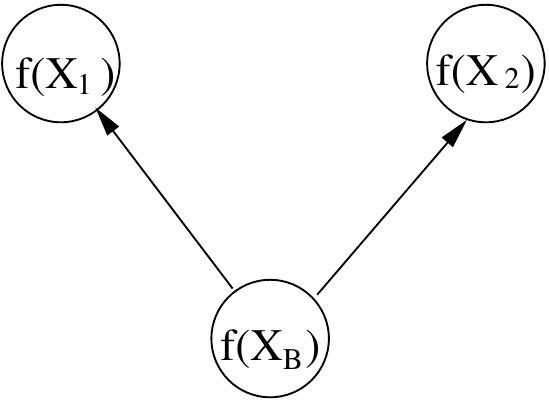}
\end{center}
\caption{\label{gm} Non-stationary GP prior graphical model.}
\end{figure}

We assume that the processes internal to each region are conditionally
independent given the process at the boundary $B$.  The corresponding
graphical model is shown in Figure~\ref{gm} and $f(X_1)$ and $f(X_2)$
are the processes internal to the regions labelled 1 and 2 and
$f(X_B)$ is the process at the boundary.
The process in region $1$ and at the boundary is modelled using the GP
kernel $K_1$.  The rows and columns of $K_1$ correspond to the stacked
vector $O_1=(X_1,X_B)^T$:
\begin{align*}
K_1=\begin{pmatrix}
K_1(X_1,X_1) & K_1(X_1,X_B)^T \\
K_1(X_1,X_B) & K_1(X_B,X_B)\
\end{pmatrix}\ .
\end{align*}
Similarly, the process in region $2$ and at the boundary is
modelled using the GP kernel $K_2$ where the row and columns correspond
to the stacked vector $O_2=(X_B,X_2)^T$:
\begin{align*}
K_2=\begin{pmatrix}
K_2(X_B,X_B) & K_2(X_2,X_B)^T\\
K_2(X_2,X_B) & K_2(X_2,X_2)
\end{pmatrix}\ .
\end{align*}
Of course, if the kernels both accurately model the prior covariance
of the process at the boundary then:
\begin{align*}
K_1(X_B,X_B)=K_2(X_B,X_B)=K_B\ .
\end{align*}
So we condition both $K_1$ and $K_2$ on $K_B$ to yield $K^*_1$ and
$K^*_2$ respectively:
\begin{align*}
K^*_1(X_1,X_2)&= K_1(X_1,X_1)+G_1[K_B-K_1(X_B,X_B)]G_1^T\ , \\
K^*_2(X_2,X_2)&= K_2(X_2,X_2)+G_2[K_B-K_2(X_B,X_B)]G_2^T
\end{align*}
where $G_1=K_1(X_1,X_B)K_1(X_B,X_B)^{-1}$ and
$G_2=K_2(X_2,X_B)K_2(X_B,X_B)^{-1}$.  The global prior covariance is
then:
\begin{align*}
K=\begin{pmatrix}
K^*_1(X_1,X_1) & K^*_1(X_1,X_B)^T &  D\\
K^*_1(X_1,X_B) & K_B & K^*_2(X_2,X_B)^T\\
D^T & K^*_2(X_2,X_B) & K^*_2(X_2,X_2) 
\end{pmatrix}\ .
\end{align*}
where the rows and columns correspond to the stacked vector
$O=(X_1,X_B,X_2)^T$.  The cross-terms, $D$, are:
\begin{align*}
D\triangleq{\rm Cov}(f^*_2(X_1),f^*_2(X_2))
\end{align*}
where $f^*_1$ and $f^*_2$ are the region function values conditioned
on the function at the boundary:
\begin{align}
f^*_1(X_1)&=K_1(X_1,X_B)K_B^{-1}f(X_B)\ ,\\
f^*_2(X_2)&=K_2(X_2,X_B)K_B^{-1}f(X_B)\ .
\end{align}
Since ${\rm Cov}(f(X_B),f(X_B))=K_B$ then:
\begin{align*}
D=G_1K_BG_2^T\ .
\end{align*}
As a corollary of this approach we can derive a Gaussian process
kernel for 1D signals with a change point at $x_B$:
\begin{corollary}
\label{cor1}
  If $K_1$ and $K_2$ are two stationary GP kernels (not necessarily
  from the same family) which model region $1$ and region $2$
  respectively, and $\theta_1$ and $\theta_2$ are their
  hyperparameters, then:
\begin{small}
\begin{align*}
K(x_1,x_2;{\theta_1,\theta_2})\hspace*{-2cm}&\\
&=
\begin{cases}
K_1(x_1,x_2;\theta_1)+g_1(x_1)[K_B-K_1(X_B,X_B;\theta_1)]g_1(x_2)^T\\
\hspace*{1cm} \text{if\ } x_1,x_2<x_B\\
K_2(x_1,x_2;\theta_2)+g_2(x_1)[K_B-K_2(X_B,X_B;\theta_2)]g_2(x_2)^T \\
\hspace*{1cm} \text{if\ } x_1,x_2\ge x_B\\
g_1(x_1)K_Bg_2(x_2)^T \\
\hspace*{1cm} \text{otherwise}\ .
\end{cases}
\end{align*}
\end{small}
where:
\begin{align*}
g_1(x_1)&=K_1(x_1,X_B;\theta_1)K_2(X_B,X_B;\theta_1)^{-1}\ ,\\
g_2(x_2)&=K_2(x_2,X_B;\theta_2)K_2(X_B,X_B;\theta_2)^{-1}.
\end{align*}
\end{corollary}

\begin{figure}[ht]
\begin{center}
\includegraphics[width=0.45\textwidth]{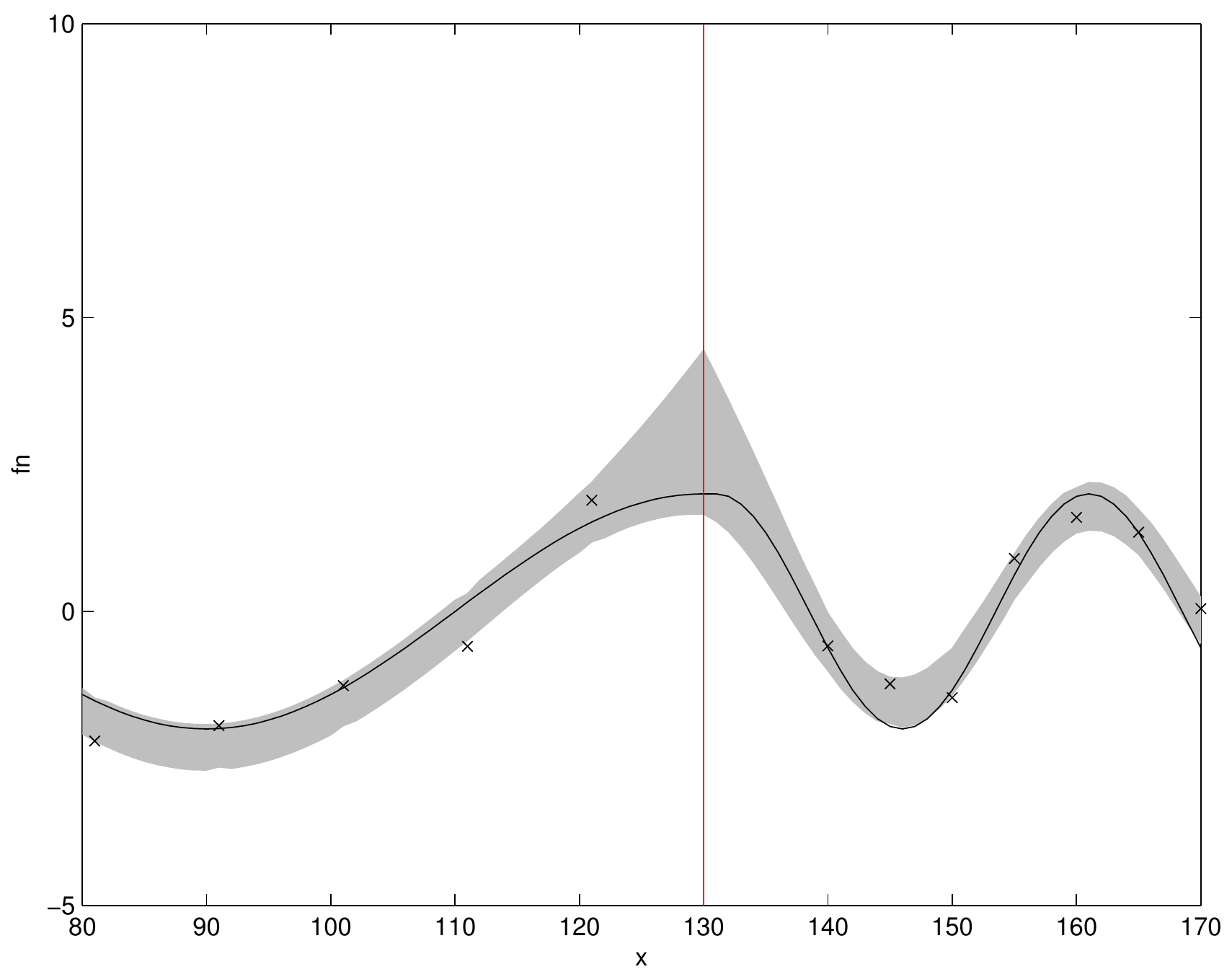}
\end{center}
\caption{\label{exgibbs4} Function and estimate obtained using
  conditionally independent function segments described by stationary
  Gibbs Gaussian process kernels joined at the boundary.}
\end{figure}

To demonstrate the new kernel we return to the problem in
Figure~\ref{exgibbs1}.  Using identical hyperparameters and
observations as in Figure~\ref{exgibbs1} the function estimate
obtained using the Markov Region Link approach is shown in
Figure~\ref{exgibbs4}.

\section{Derivative Boundary Conditions}
\label{sec:derivbnd}
So far, we have developed covariance functions which preserve function
continuity at the boundary.  The approach can be extended to assert
function derivative continuity at the boundary.  The covariance
between a function and any of its derivatives can be determined from
the GP kernel~\cite{rasmussen06}.  For example, the prior covariance
between the function and its first derivative is:
\begin{align*}
  [\partial K(X,Y)]_{ij}&\triangleq {\rm Cov}\left(\frac{\partial f(x_i)}{\partial
      x_{i}},f(x_j)\right)\\
&=\frac{\partial K(x_i,x_j)}{\partial x_{i}}\ .
\end{align*}
where $x_i\in X$ and $x_j\in Y$. The covariance between the
derivatives is:
\begin{align}
  [\partial\partial K(X,Y)]_{ij}&\triangleq
  {\rm Cov}\left(\frac{\partial f(x_i)}{\partial x_{i}},\frac{\partial
      f(x_j)}{\partial x_j}\right)\nonumber\\&=\frac{\partial^2
    K(x_i,x_j)}{\partial x_{i}\partial x_j}\ .
\label{dd}
\end{align}
The derivative variance at $x_i$ can be obtained by setting $x_j$ to
$x_i$ in Eqn~\ref{dd}.  In our notation $\partial K(X,Y)$ denotes
partial derivative with respect to the first parameter, in this
case $X$ and $\partial\partial K(X,Y)$ denotes double differentiation
with respect to both $X$ and then $Y$.

These relationships can be used to define non-stationary GP covariance
priors which impose continuous function derivatives at region
boundaries.  The prior mean and covariance for both the regional and
global priors are augmented to include the function derivative.  For
example, if the first derivative is added to the prior then the prior
covariances for regions $R_1$ and $R_2$ become:~\footnote{We shall use
  prime to denote the augmented covariances.}

\begin{small}
\begin{align*}
K'_1=\begin{pmatrix}
K_1(X_1,X_1) & K_1(X_1,X_B) & [\partial K_1(X_B,X_1)]^T\\
K_1(X_B,X_1) & K_1(X_B,X_B) & [\partial K_1(X_B,X_B)]^T\\
\partial K_1(X_B,X_1) & \partial K_1(X_B,X_B) & \partial\partial K_1(X_B,X_B)
\end{pmatrix}
\end{align*}
\end{small}

and:

\begin{small}
\begin{align*}
K'_2=\begin{pmatrix}
K_2(X_B,X_B) & [\partial K_2(X_B,X_B)]^T & K_2(X_B,X_2)\\
\partial K_2(X_B,X_B) & \partial\partial K_2(X_B,X_B) & \partial K_2(X_B,X_2)\\
[\partial K_2(X_B,X_2)]^T & [\partial K(X_B,X_2)]^T & K_2(X_2,X_2)
\end{pmatrix}
\end{align*}
\end{small}

The rows and columns in $K'_1$ correspond to the stacked vector
$O_1=(X_1,X_B,d(X_B))^T$ where $d(X_B)$ denotes the function
derivative at $X_B$.  Similarly, the rows and columns in $K'_2$
correspond to the stacked vector $O_2=(X_B,d(X_B),X_2)$.
Consequently, we can use the approach outlined in
Section~\ref{sec:MRL} to construct a global prior for processes which
are conditionally independent in each region.  This is done by
defining $K_B$ as follows:
\begin{align*}
K_B=
\begin{pmatrix}
 K_1(X_B,X_B) & [\partial K_1(X_B,X_B)]^T\\
\partial K_1(X_B,X_B) & \partial\partial K_1(X_B,X_B)
\end{pmatrix}
\end{align*}
and using $K'_1$ and $K'_2$ defined above in place of $K_1$ and $K_2$
in Section~\ref{sec:MRL}.

\begin{figure}[ht]
\begin{center}
\includegraphics[width=0.45\textwidth]{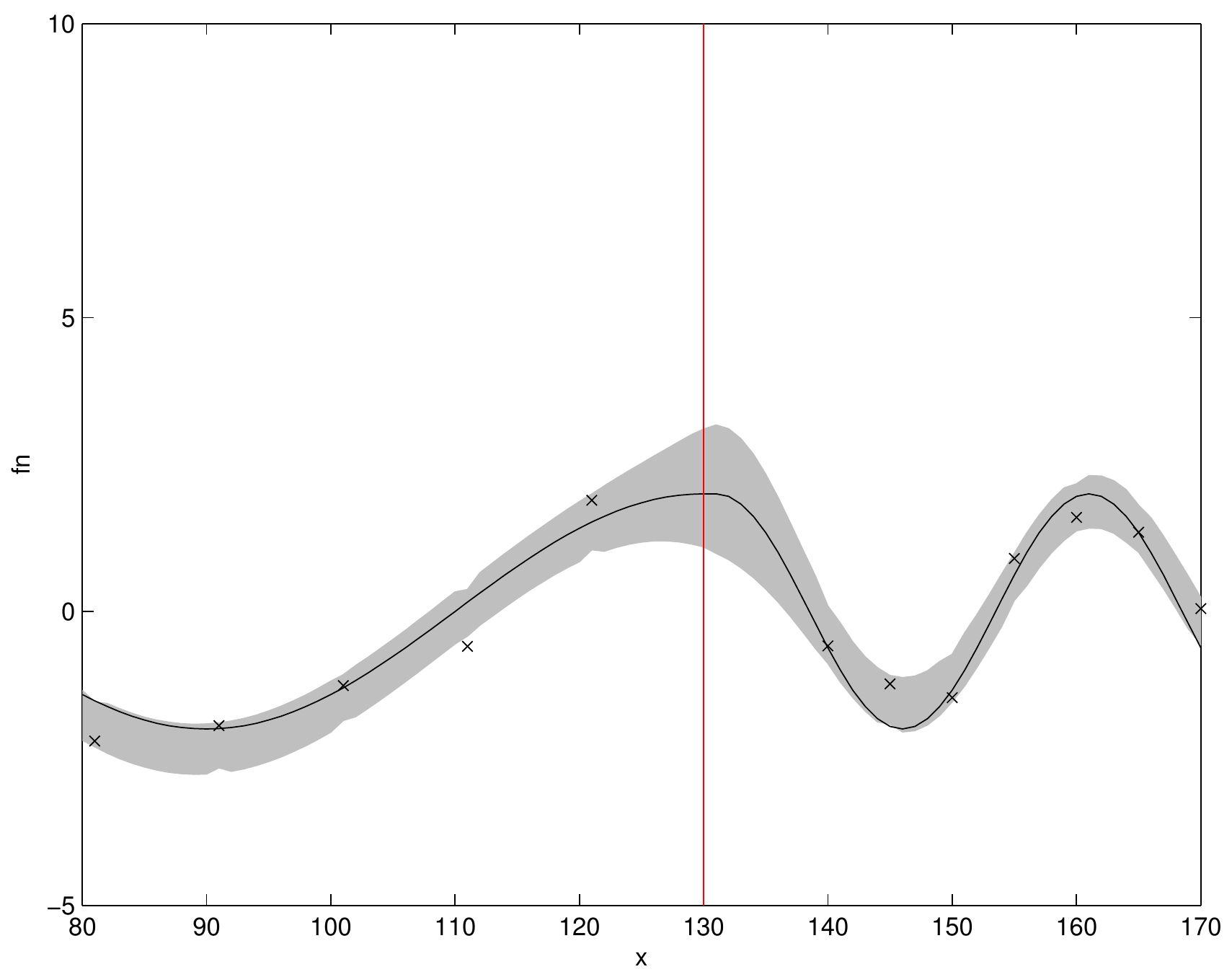}
\end{center}
\caption{\label{exgibbs5} Function estimated using two stationary
  Gibbs kernels joined at $x=130$ with the constraint that the
  function first derivative is continuous at $x=130$.}
\end{figure}

Figure~\ref{exgibbs5} shows the effect that the derivative constraint
can have on the function estimate.  Two stationary Gibbs kernels are
used with $l(x)=35$ for $x\le 130$ and $l(x)=15$ for $x>130$ as in
Figure~\ref{exgibbs1}.  Clearly, the approach which imposes a
continuous first derivative on the GP model produces a tighter
function estimate at $x=130$.

\section{Application 1: Target Estimation with Faulty Sensors}
\label{sec:appl1}
We shall use a GP to estimate a target's position over time $t$ as it
is tracked by a simple sensor.  However, the sensor is faulty and
outputs readings which have either drifted gradually from the truth
or undergone a sudden jolt away from the truth resulting in a fixed bias
in the reading (see Figure~\ref{fault1}).

The proposed filter algorithm operates on-line and infers a posterior
distribution for the current target's position using observations of
its previous locations.  Smoothing from future observations is not
considered.  The target's trajectory is described by the process $f$
and the sensor is subject to occasional faults.  The sensor's fault
process, $e$, can be either a short term fixed bias or it can drift
over a period of time.  The, possibly faulty, observation at time
$t_i$ is:
\begin{align*}
y_i=f(t_i)+e(t_i)+\epsilon_i
\end{align*}
where $\epsilon_i$ is zero mean, Gaussian with variance
$\sigma^2=0.001$.  We wish to estimate target location $f(t)$ over
time.  

The processes $f$ and $e$ are modelled by GP kernels $K_f$ and $K_e$
respectively.  We will assume that $K_f$ is stationary and we will use
a simple squared exponential kernel to model the target dynamics.
However, the fault is intermittent and it starts at time $t=T_0$ and
ends at $t=T_1$.  We model the fault process using a non-stationary
kernel.  Firstly, $e(t)$ is zero over times, $t<T_0$ and $t>T_1$, for
which there is no fault.  For a bias fault, we assume that the bias is
a fixed offset and thus assert:
\begin{align*}
K_{\text bias}(t_i,t_j)=\mu\ \text{for all\ } T_0\le t_i,t_j\le T_1\ .
\end{align*}
where $\mu$ is a scale parameter representing the magnitude of the
bias.  We assume that the drift is gradual and build the drift model using a
squared exponential kernel:
\begin{align*}
K_{\text drift}(t_i,t_j)=\mu \exp\left(-\frac{(t_i-t_j)^2}{L^2}\right)
\end{align*} 
that will be modified shortly to accommodate drift error starting from
zero.  The parameters $\mu$ and $L$ are the output and input scales,
respectively, and again, $T_0\le t_i,t_j\le T_1$. 

The bias fault causes a jump in the observation sequence when the
fault sets in at $t=T_0$.  However, the drift fault is gradual and
$e(T)$ is zero at $t=T_0$ and discontinuous at $T_1$ when the fault
disappears (see Figure~\ref{fault1}). We use the Markov Region Link
kernel to construct the drift fault process prior covariance
$K^*_\text{drift}$ from $K_\text{drift}$ and impose the continuity
boundary condition at $T_0$.  Using the approach set out in
Section~\ref{sec:MRL} the prior covariance for the drift fault becomes
the block matrix:
\begin{align*}
K^*_\text{drift}=\begin{pmatrix}
0 & 0 &0 & 0\\
0 &0 & K^*_\text{drift}(T_0,X_f)^T& 0\\
0 & K^*_\text{drift}(X_f,T_0) & K^*_\text{drift}(X_f,X_f) & 0\\
0 & 0 & 0 & 0
\end{pmatrix}\ .
\end{align*}
The first row and column are zero matrices for times less than $T_0$,
corresponding to the period before the fault starts.  The last row and
column are zero matrices for times greater than $T_1$, corresponding
to times after the fault has stopped.  The central row and column
blocks are prior covariances over time samples $X_f$ during which the
sensor is faulty:
\begin{align*}
X_f&=\{t\mid T_0< t \le T_1\}\ .
\end{align*}
Continuity of the fault process at $T_0$ imposes
$K^*_\text{drift}(T_0,T_0)=0$.  Values for $K^*_\text{drift}$
are obtained using Corollary~\ref{cor1} in Section~\ref{sec:MRL} with
$X_B=T_0$, $K_1=0$, $K_B=K^*_\text{drift}(T_0,T_0)=0$ and
$K_2=K_\text{drift}$.

The bias kernel is more straight forward:
\begin{align*}
K_\text{bias}=\begin{pmatrix}
0 & 0 &0 & 0\\
0 &\mu & \mu& 0\\
0 & \mu & \mu & 0\\
0 & 0 & 0 & 0
\end{pmatrix}
\end{align*}
with the rows and columns interpreted as for $K_\text{drift}$.

\begin{figure}[ht]
\begin{center}
\begin{tabular}{cc}
\includegraphics[width=0.2\textwidth]{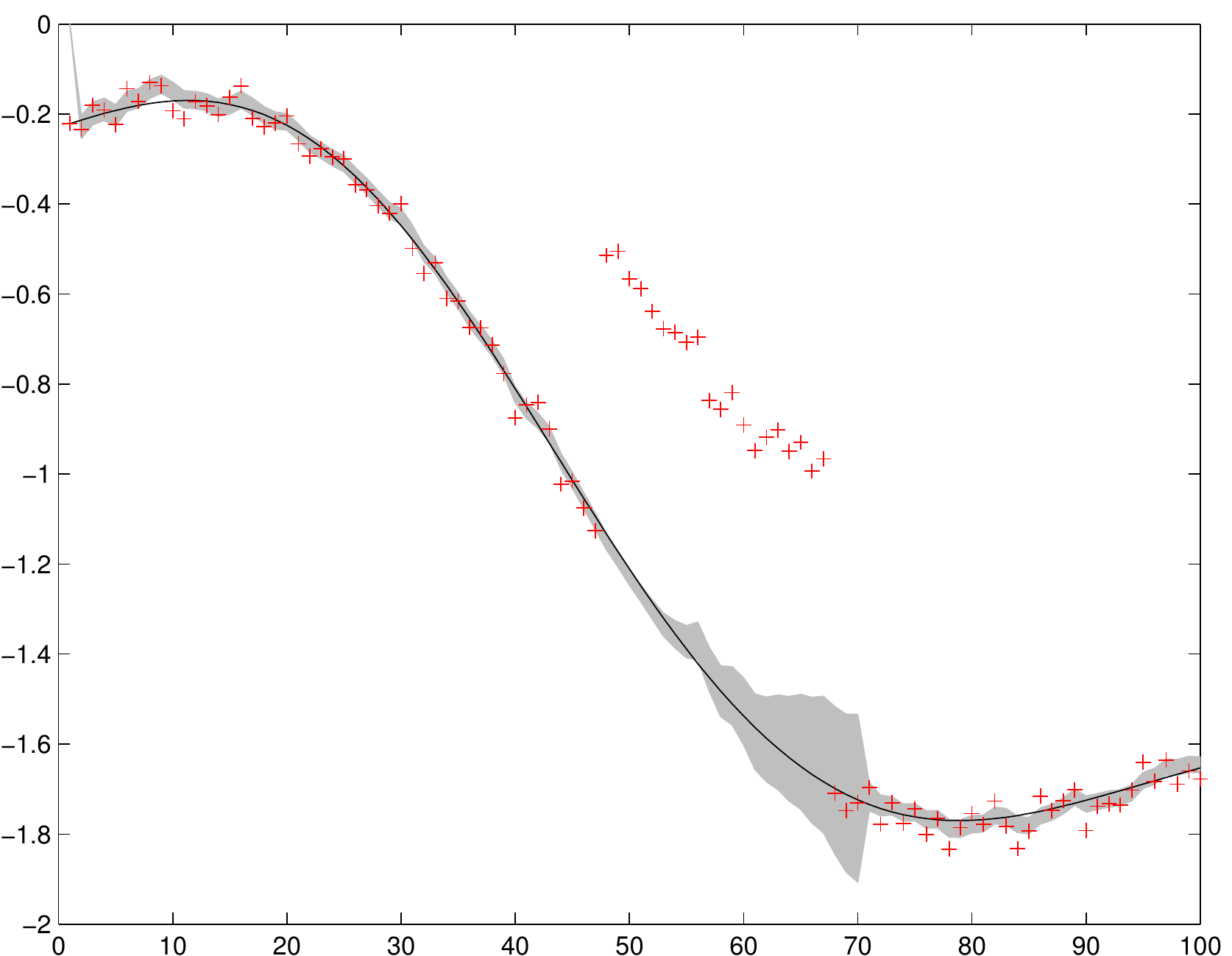}&
\includegraphics[width=0.2\textwidth]{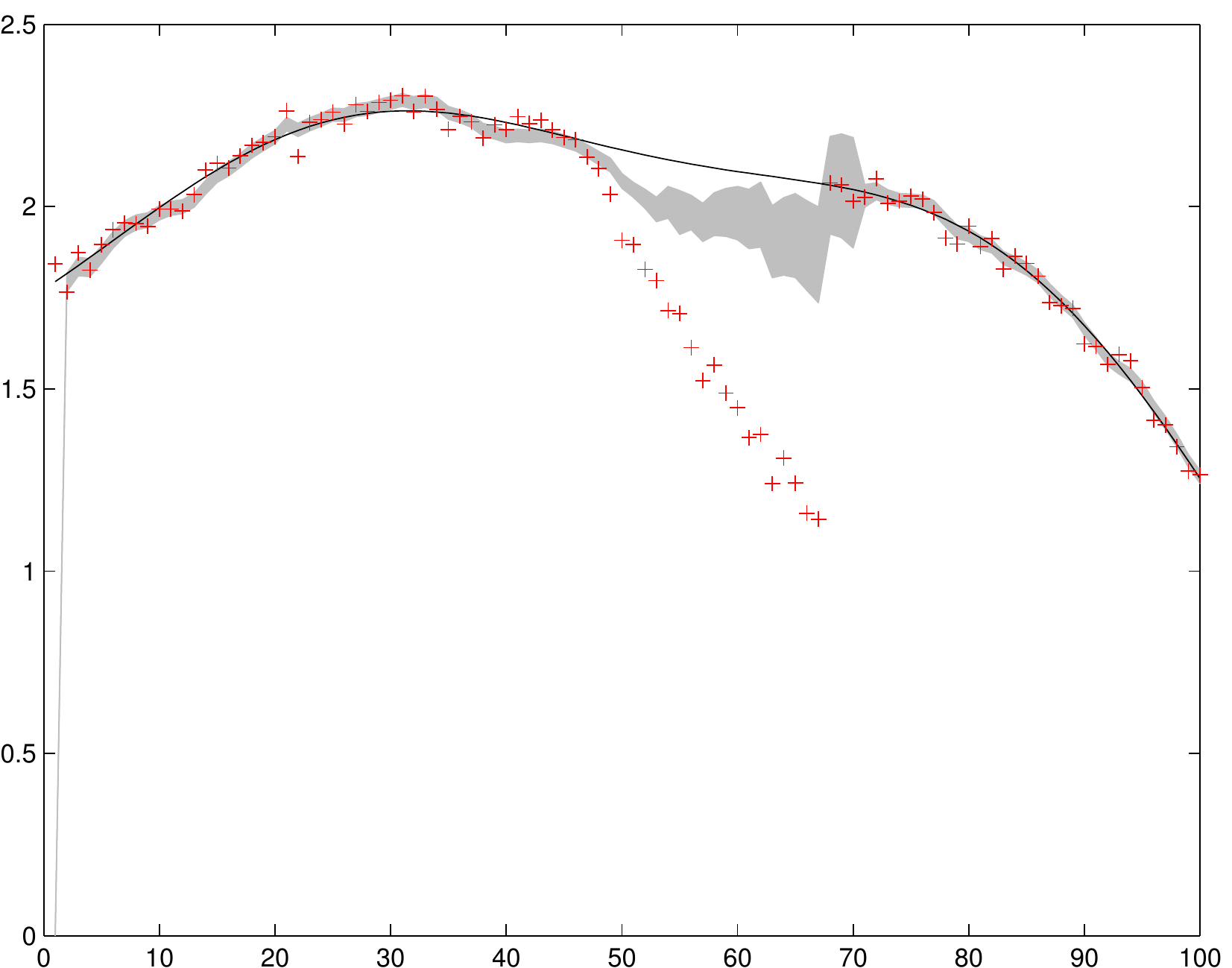}\\
 (a) Bias & (b) Drift
\end{tabular}
\end{center}
\caption{\label{fault2} Typical target trajectory and faulty observations
for both bias and drift faults.}
\end{figure}

Figure~\ref{fault2} shows typical tracks, observations and GP track
estimates.  The target trajectory and observation sequence are
randomly generated, $f\sim\Bbb{N}(0,K_f)$ and
$y\sim\Bbb{N}(0,K_e+\sigma^2I)$.  Notice that the algorithm has
successfully corrected for the faulty observations.

Parallel faults may also be modelled using our approach.  A parallel
fault comprises more than one basic fault occurring at the same
time.  In the following example the sensor measurement drifts to a 
new fixed bias offset.  The fault can be modelled using a drift fault,
without fault correction, followed immediately by a biased fault.

\begin{figure}[ht]
\begin{center}
\includegraphics[width=0.5\textwidth]{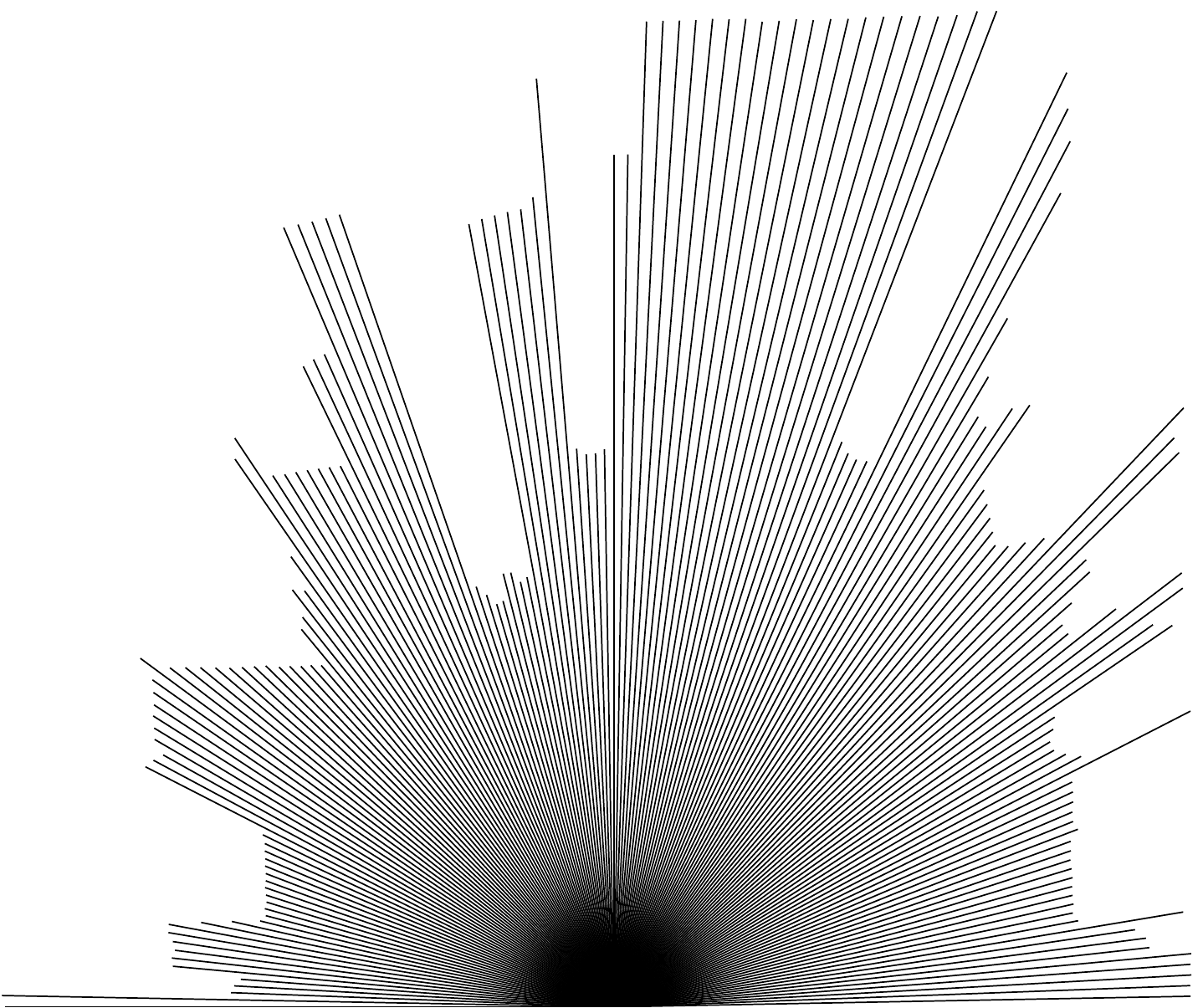}
\end{center}
\caption{\label{scan_example} Typical SICK sensor range/bearing scan of the laboratory.}
\end{figure}

\begin{figure}[ht]
\begin{center}
\includegraphics[width=0.3\textwidth]{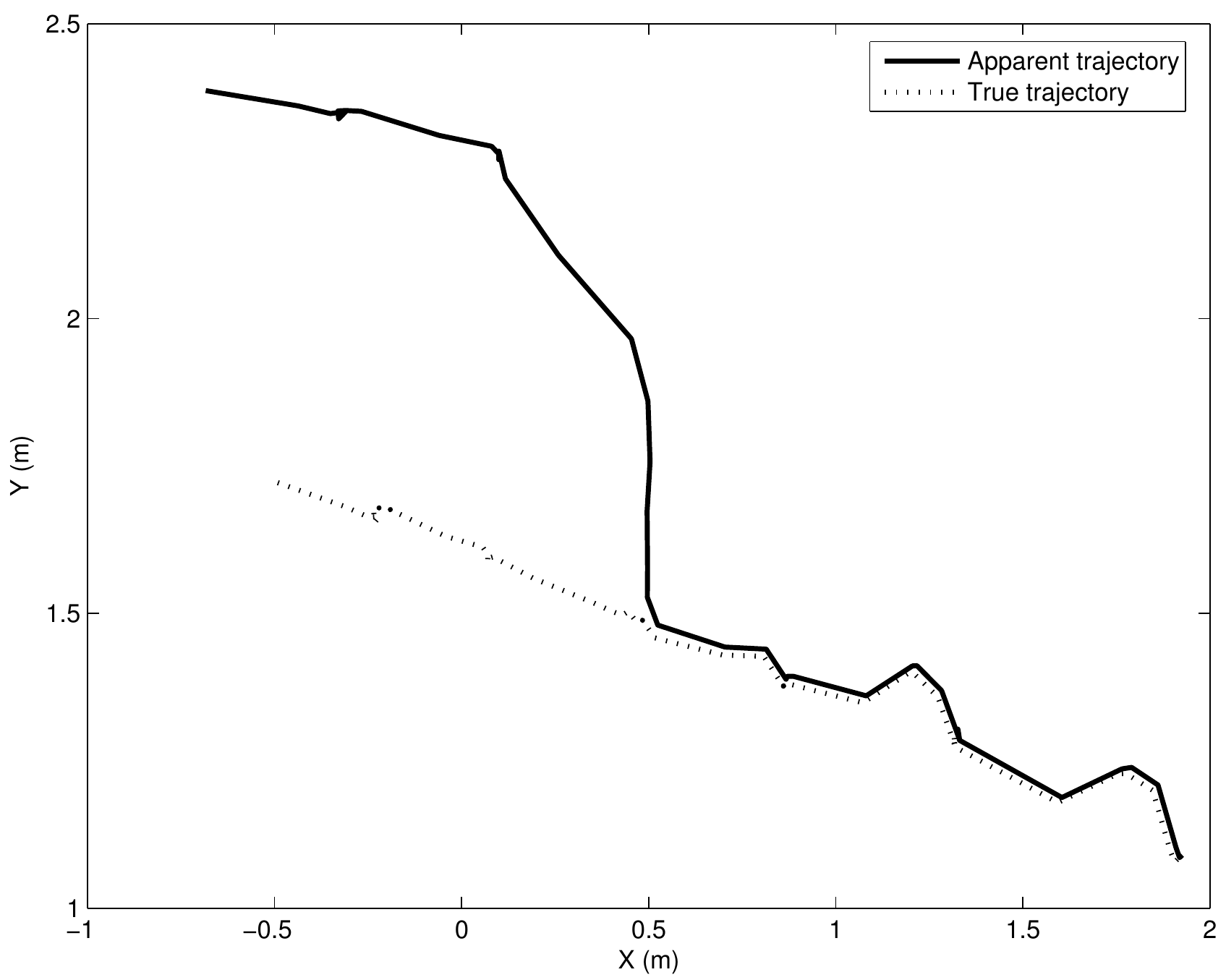}
\end{center}
\caption{\label{true_traj} True and observed target trajectory
  obtained using SICK range sensor.}
\end{figure}

A SICK laser range sensor is used to track a person.  The sensor is
subject to a knock as it tracks the person.  This knock results in the
sensor drifting for a period before coming to rest.  The sensor data
thus exhibits a drift followed by a constant bias.
Figure~\ref{scan_example} shows a typical $180$ bearing range scan at a
single time instance during the tracking procedure.
Figure~\ref{true_traj} shows the data obtained from our sensor
(transformed to the laboratory centred Cartesian coordinates) and also
the target's true trajectory obtained by using objects in the
environment with known location.

We use the MRL kernel to splice together both the drift kernel and the
bias kernel and assert that the target observations are continuous at
the transition from drift to bias fault.  Both the transition time and
the kernel induced variance, $K_B$, at that time are parameters
of our model.

\begin{figure}[ht]
\begin{center}
\includegraphics[width=0.5\textwidth]{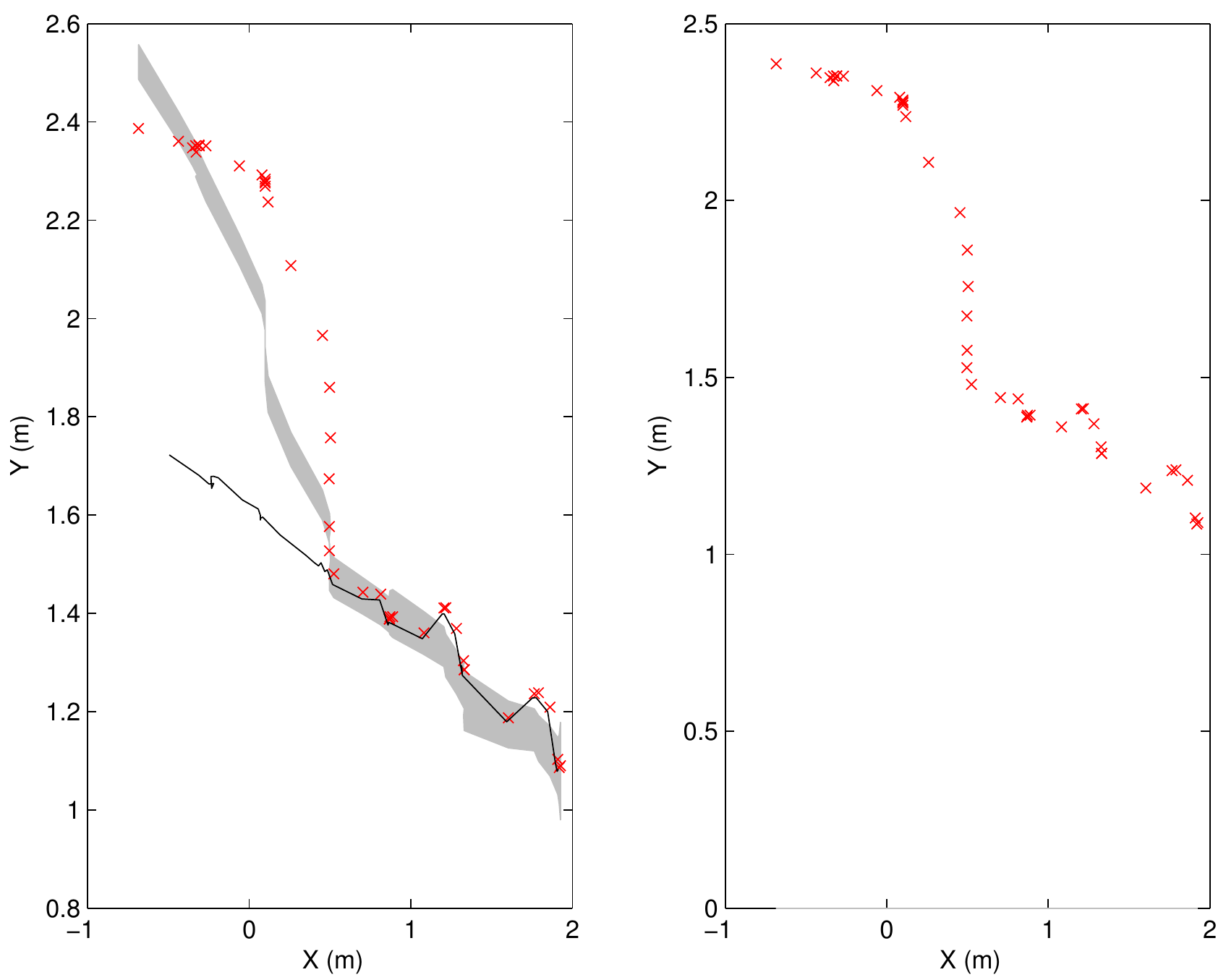}
\end{center}
\caption{\label{nofault} GP target trajectory estimate obtained by ignoring fault.}
\end{figure}

Figure~\ref{nofault} shows the ground truth and the one standard deviation
estimate if no fault is assumed as the person moves from right to
left.  The magnitude of the error is under-estimated.

\begin{figure}[ht]
\begin{center}
\includegraphics[width=0.5\textwidth]{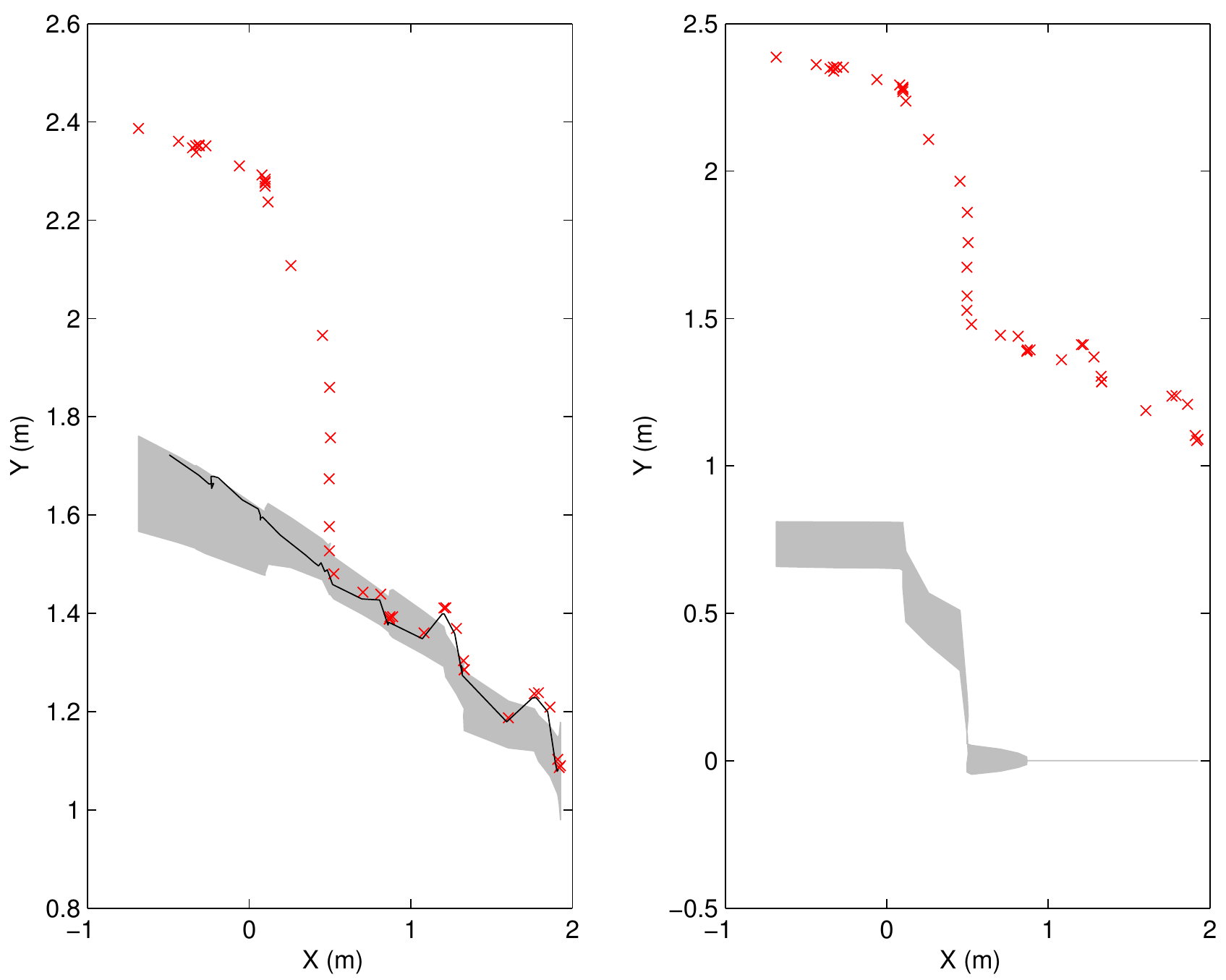}
\end{center}
\caption{\label{traj_est} GP target trajectory estimate obtained after
  recovering from faults.}
\end{figure}

Figure~\ref{traj_est} shows the GP on-line estimated trajectory as well
as an estimate of the fault.  Both estimates are correctly determined
in this case.

\section{Application 2: EOG Artifact Removal From EEG Signals}
\label{sec:appl2}
In this example we detect and remove EOG artifacts from EEG signals by
modelling the artifacts as drift type faults. The recovery of the EEG
signal is often treated as a blind source separation
problem~\cite{roberts99} where ICA identifies the separate artifact
free EEG signal (which we refer to as EEG*) and the EOG signal.  In
our approach we use explicit models for the EEG* and EOG signal
and stipulate that these signals are independent.

The ``observed'' EEG signal, $y$, is a mixture of EEG* and EOG
artifact signals.  The EEG* signal, $s_{eeg*}$, is modelled as the
combination of a smooth function $m_{eeg*}$ (generated by a GP with
prior covariance $K_{eeg*}$) and i.i.d distributed {\em residuals}
$r_{eeg*}$.  The EOG artifact, $s_{eog}$, is modelled as a piece-wise
smooth function, $m_{eog}$ (generated from a GP ``fault'' model with
prior covariance $K_{eog}$) and, similarly, for the residuals $r_{eog}$.
\begin{align*}
s_{eeg*}&= m_{eeg*}+r_{eeg*}\ ,\\
s_{eog}&= m_{eog}+r_{eog}\ ,\\
y&=s_{eeg*}+s_{eog}\ .
\end{align*}
where $m_{eeg*}\sim \Bbb{N}(0,K_{eeg*})$,
$r_{eeg*}\sim \Bbb{N}(0,R_{eeg*}I)$, $m_{eog}\sim \Bbb{N}(0,K_{eog})$
and $r_{eog}\sim \Bbb{N}(0,R_{eog}I)$ and $I$ is the identity matrix.
The random vectors $m_{eeg*}$, $m_{eog}$, $r_{eeg*}$ and $r_{eog}$ are
assumed to be mutually independent.  The residuals, $r_{eeg*}$ and
$r_{eog}$, are considered to be part of the signals and are therefore
not treated as noise and are not filtered out.

We use a simple squared exponential to model the EEG* signal.
Typically the EOG signal is zero everywhere except within a small time
window.  Within this window the EOG artifact can be modelled as two
smooth functions (not necessarily monotonic) which join at a spike
near the centre of the window (so that the EOG signal is continuous
but not differentiable at the join).  Thus, the EOG's prior covariance
$K_{eog}$ is chosen to be zero everywhere except between the artifacts
start and end times, $T_s$ and $T_e$, respectively.  We use the
methods outlined in Section~\ref{sec:MRL} to build the EOG artifact
prior covariance.  Between the start and end times the EOG artifact
signal is modelled by two piece-wise squared exponential kernels
joined mid-way between $T_s$ and $T_e$ so that they are continuous at
the mid-point.  Furthermore, the EOG signal is zero at $T_s$ and
$T_e$.

The following GP equations determine the mean and covariance for the
hidden variable, $m_{eeg*}$:
\begin{small}
\begin{align*}
\hat{m}_{eeg*}(x_*)&=[K_{eeg*}(x_*,X)]\times\\
&\hspace*{1cm}[K_{eeg*}(X,X)+K_{eog}(X,X)+\sigma^2I]^{-1} y(X)\\
{\rm Cov}_{eeg*}(x_*)&= K_{eeg*}(x_*,x_*)-\\
&\hspace*{1cm}K_{eeg*}(x_*,X)\times\\
&\hspace*{1cm}[K_{eeg*}(X,X)+K_{eog}(X,X)+\sigma^2I]^{-1}\times\\
&\hspace*{1cm}K_{eeg*}(x_*,X)^T
\end{align*}
\end{small}
where $\sigma^2=R_{eeg*}+R_{eog}$. Similar expressions can be obtained
for $m_{eog}$.

To track the EEG signal our algorithm determines $s_{eeg*}$
sequentially over time.  When $x_*$ is the current time and $X$ is the
previous times at which data points were obtained then:
\begin{small}
\begin{align*}
p(s_{eeg*}(x_*),s_{eog}(x_*)\mid y(x_*),\hat{m}_{eeg*}(x_*),\hat{m}_{eog}(x_*))\hspace*{-6cm}&\\
 &\propto p(y(x_*)\mid s_{eeg*}(x_*),s_{eog}(x_*),\hat{m}_{eeg*}(x_*),\hat{m}_{eog}(x_*))\\
&\ \ \ \ \times p(s_{eeg*}(x_*),s_{eog}(x_*)\mid \hat{m}_{eeg*}(x_*),\hat{m}_{eog}(x_*))\\
&\propto \delta_{y(x_*),s_{eeg*}(x_*)+s_{eog}(x_*)}\\
&\ \ \ \ \times p(s_{eog}(x_*)\mid \hat{m}_{eog}(x_*))\\
&\ \ \ \ \times p(s_{eeg*}(x_*)\mid \hat{m}_{eeg*}(x_*))\ .
\end{align*}
\end{small}
Marginalising $s_{eog}$:
\begin{align*}
p(s_{eeg*}(x_*)\mid y(x_*),\hat{m}_{eeg*}(x_*),\hat{m}_{eog}(x_*))\hspace*{-4cm}&\\
& \propto p(y(x_*)-s_{eeg*}(x_*)\mid \hat{m}_{eog}(x_*))\\
&\ \ \ \ \times p(s_{eeg*}(x_*)\mid \hat{m}_{eeg*}(x_*))\ .
\end{align*}
In general, when $s$ is Gaussian distributed then its mean, $\hat{s}$, is the
solution to:
\begin{align*}
\frac{\partial \log p(s\mid \cdot)}{\partial s}=0
\end{align*}
and its variance, $\text{Var}(s)$, is given by:
\begin{align*}
\text{Var}(s)&=-E\left[\frac{\partial^2 \log p(s\mid \cdot)}{\partial s^2}\right]^{-1}\ .
\end{align*}
Thus, defining $P^*_{eeg*}(x_*)\triangleq\text{Cov}_{eeg*}(x_*)+R_{eeg*}$ and $P^*_{eog}(x_*)\triangleq\text{Cov}_{eog}(x_*)+R_{eog}$:
\begin{align*}
\hat{s}_{eeg*}(x_*)=\frac{P^*_{eeg*}(x_*)(y(x_*)-\hat{m}_{eog}(x_*))+P^*_{eog}(x_*)\hat{m}_{eeg*}(x_*)}{P^*_{eeg*}(x_*)+P^*_{eog}(x_*)}
\end{align*}
and:
\begin{align*}
\text{Var}_{eeg*}(x_*)=\frac{P^*_{eeg*}(x_*)+P^*_{eog}(x_*)}{P^*_{eeg*}(x_*) P^*_{eog}(x_*)}\ .
\end{align*}
Similar reasoning leads us to similar expressions for the EOG artifact signal:
\begin{align*}
\hat{s}_{eog}(x_*)&=\frac{P^*_{eog}(x_*)(y(x_*)-\hat{m}_{eeg*}(x_*))+P^*_{eeg*}(x_*)\hat{m}_{eog}(x_*)}{P^*_{eeg*}(x_*)+P^*_{eog}(x_*)}
\end{align*}
and:
\begin{align*}
\text{Var}_{eog}(x_*)&=\frac{P^*_{eeg*}(x_*)+P^*_{eog}(x_*)}{P^*_{eeg*}(x_*) P^*_{eog}(x_*)}\ .
\end{align*}
These expressions for $\hat{s}_{eeg*}$ and $\hat{s}_{eog}$ determine
the proportion of the EEG signal residual that is assigned to the EOG
artifact signal and also to the artifact free EEG signal (EEG*).

Our model requires eight hyperparameters, collectively referred to as
$\theta$: the scale heights and scale lengths for the GP models (we
assume that both parts of the EOG model have the same scale heights
and lengths); the artifact start and end times and also the residual
variances $R_{eeg*}$ and $R_{eog}$.  The likelihood used in
Equation~\ref{learning} to determine a distribution over the
hyperparameter values is given by:
\begin{align*}
p(y(x_*)\mid \theta)&=\Bbb{N}[y(x_*); \hat{m}_{eeg*,\theta}(x_*)+\hat{m}_{eog,\theta}(x_*),\\
&\hspace*{2cm}P^*_{eeg,\theta}(x_*)+P^*_{eog,\theta}(x_*)]\ . 
\end{align*}
The hyperparameters are marginalised using Monte-Carlo
sampling.

\commentout{We initially demonstrate the efficacy of our algorithm on artificial
data so that we have some knowledge of the ground truth.  Then we
apply our algorithm on real EEG data.

\begin{figure}[ht]
\begin{center}
\includegraphics[width=0.45\textwidth]{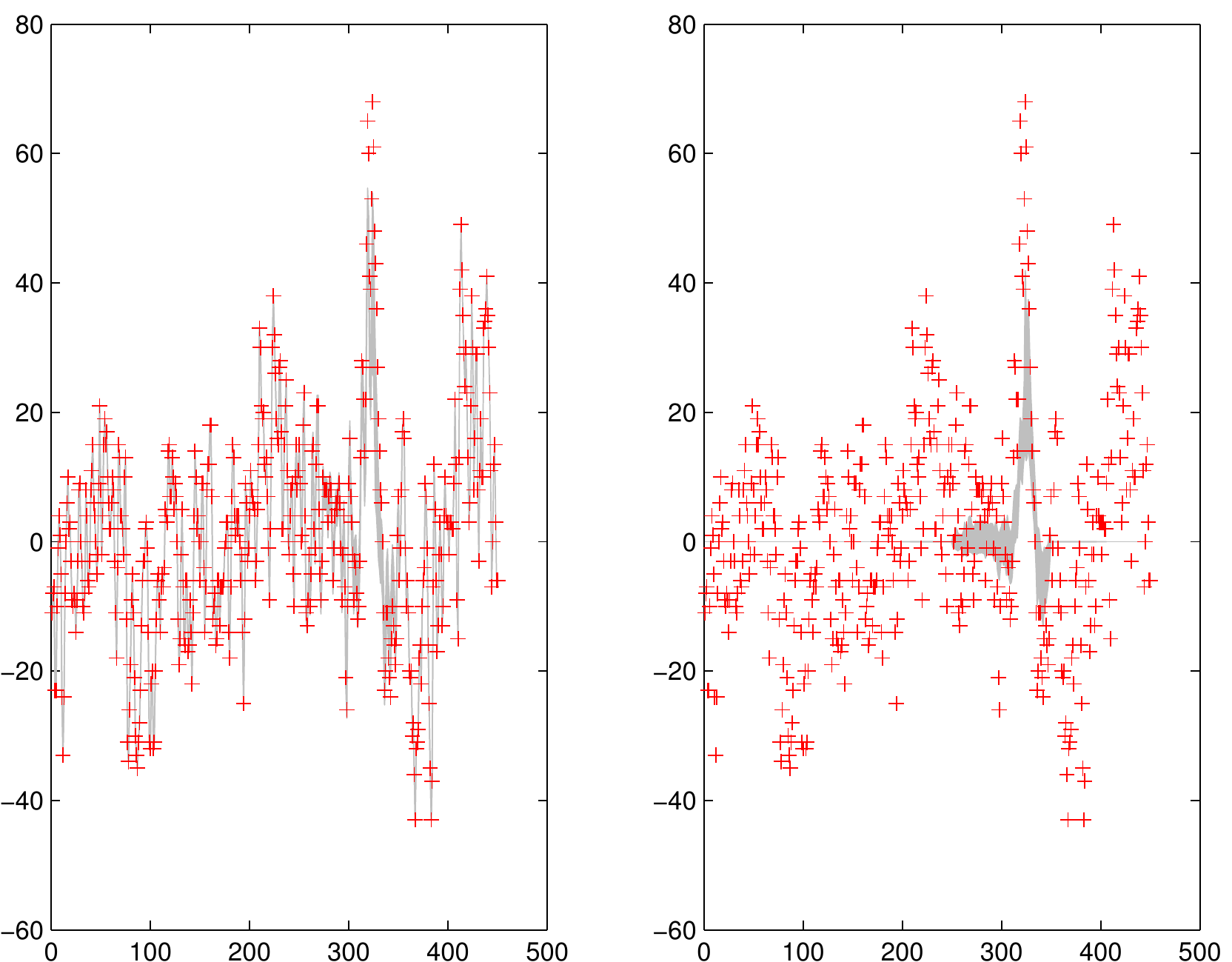}
\end{center}
\caption{\label{eeg_real} Simulated EEG signal (crosses) and one
  standard error confidence intervals for the EEG* (left panel) and
  EOG (right panel) signals obtained using the GP mixture model
  approach.}
\end{figure}

\begin{figure}[ht]
\begin{center}
\includegraphics[width=0.45\textwidth]{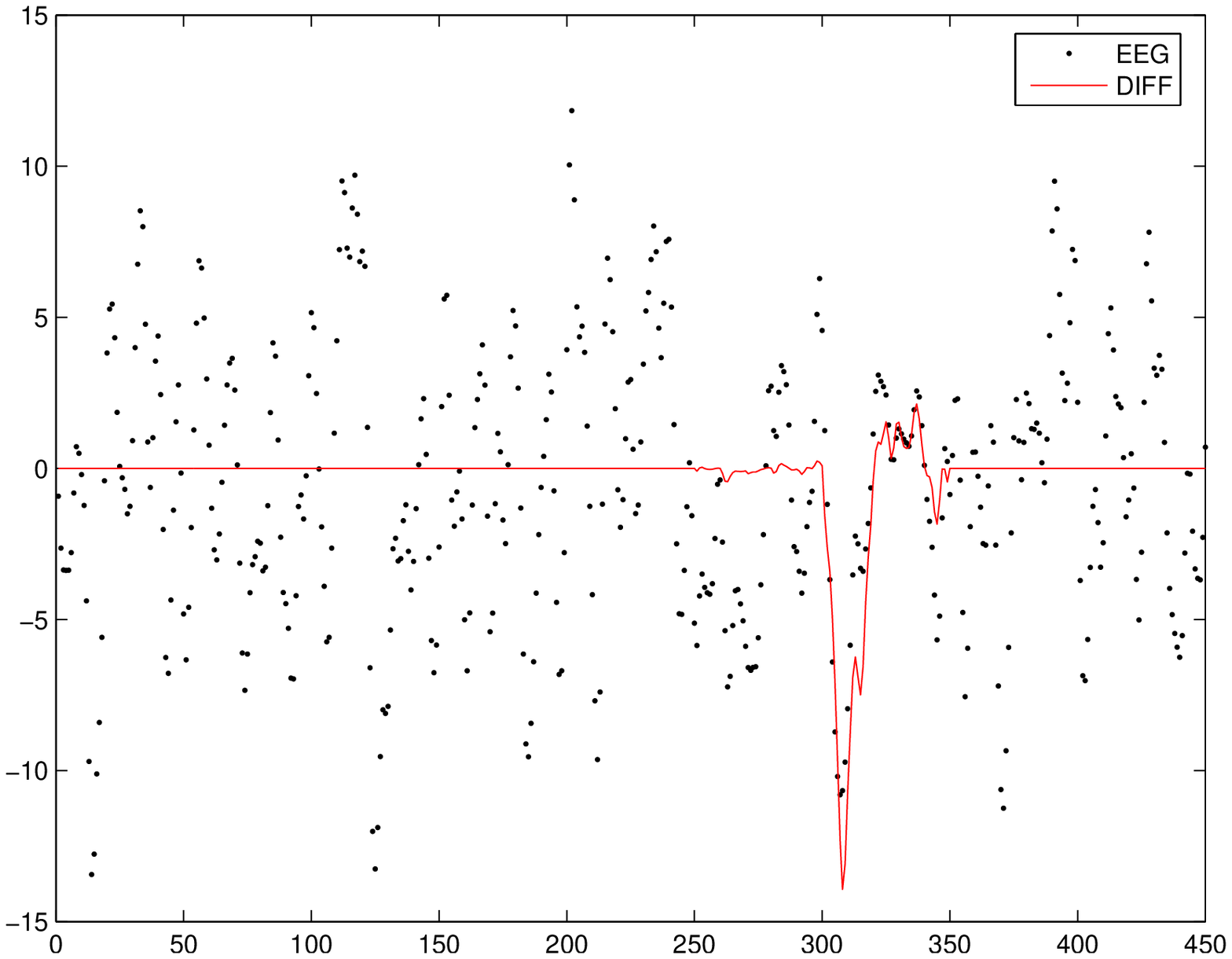}
\end{center}
\caption{\label{sigdecomp_real} Simulated EEG signal (dots) and
  difference (line) between original signal and the mean EEG* obtained
  using the GP mixture model approach.}
\end{figure}

Figure~\ref{eeg_sim} shows a simulated EEG signal which is corrupted
by EOG artifacts.  It also shows the one standard error confidence
intervals for the artifact free EEG* signal and the EOG artifact
obtained using our algorithm.  Figure~\ref{sigdecomp_real} shows the
mean difference between the original EEG signal and the inferred EEG*
signal, indicating the expected proportion of the original signal that
is retained in the EEG*.}

\begin{figure}[ht]
\begin{center}
\includegraphics[width=0.45\textwidth]{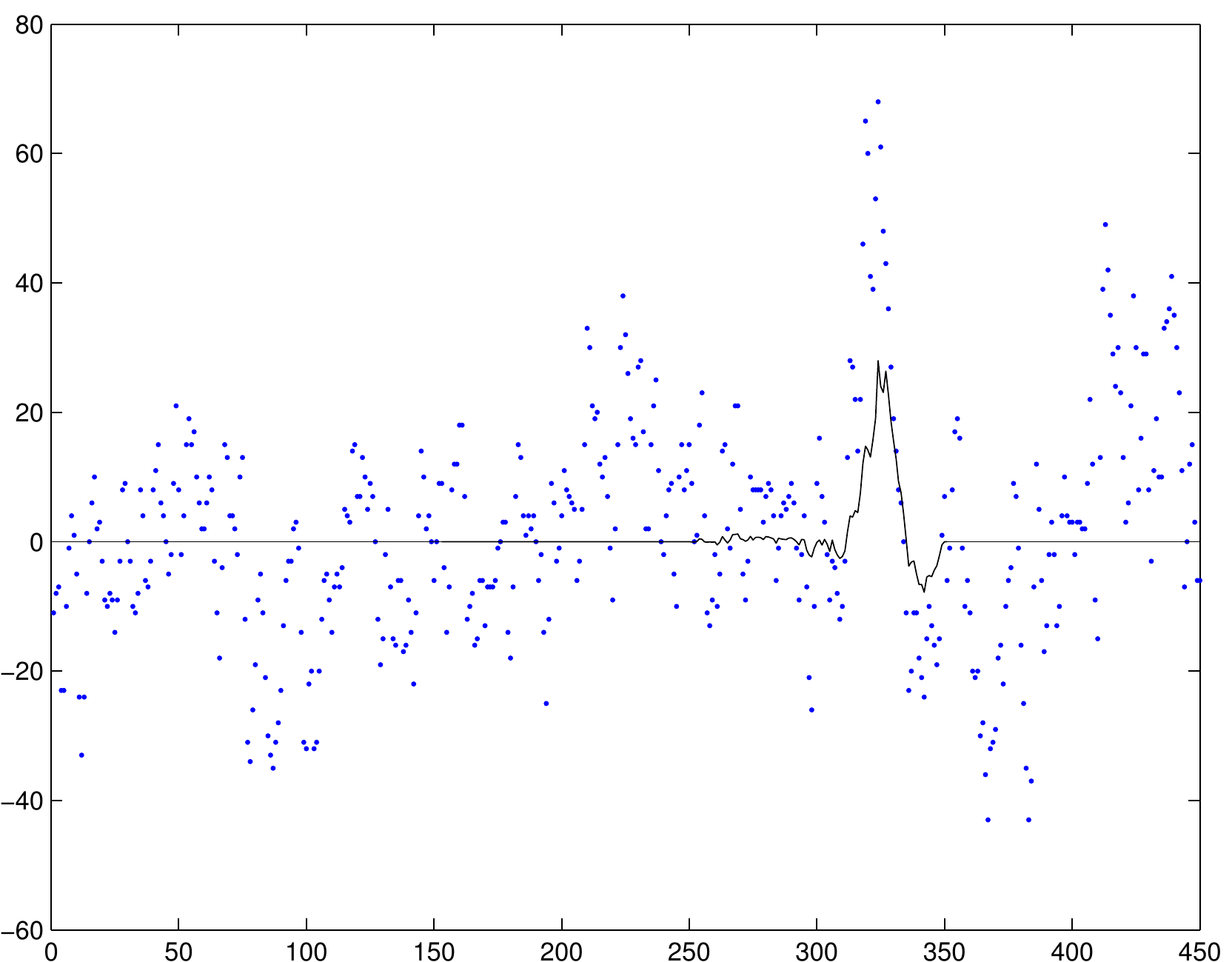}
\end{center}
\caption{\label{diff_real} Original EEG signal (dots) and difference (line) between original signal and the mean EEG* obtained using the GP mixture model approach.}
\end{figure}

Figure~\ref{eeg_real} shows a typical EEG signal which is corrupted by
EOG artifacts.  It also shows the one standard error confidence interval
for the artifact free EEG* signal and the EOG artifact obtained using
our algorithm.  Figure~\ref{diff_real} shows the mean difference
between the original EEG signal and the inferred EEG* signal,
indicating the expected proportion of the original signal that is
retained in the EEG*.
\begin{figure}[ht]
\begin{center}
\includegraphics[width=0.45\textwidth]{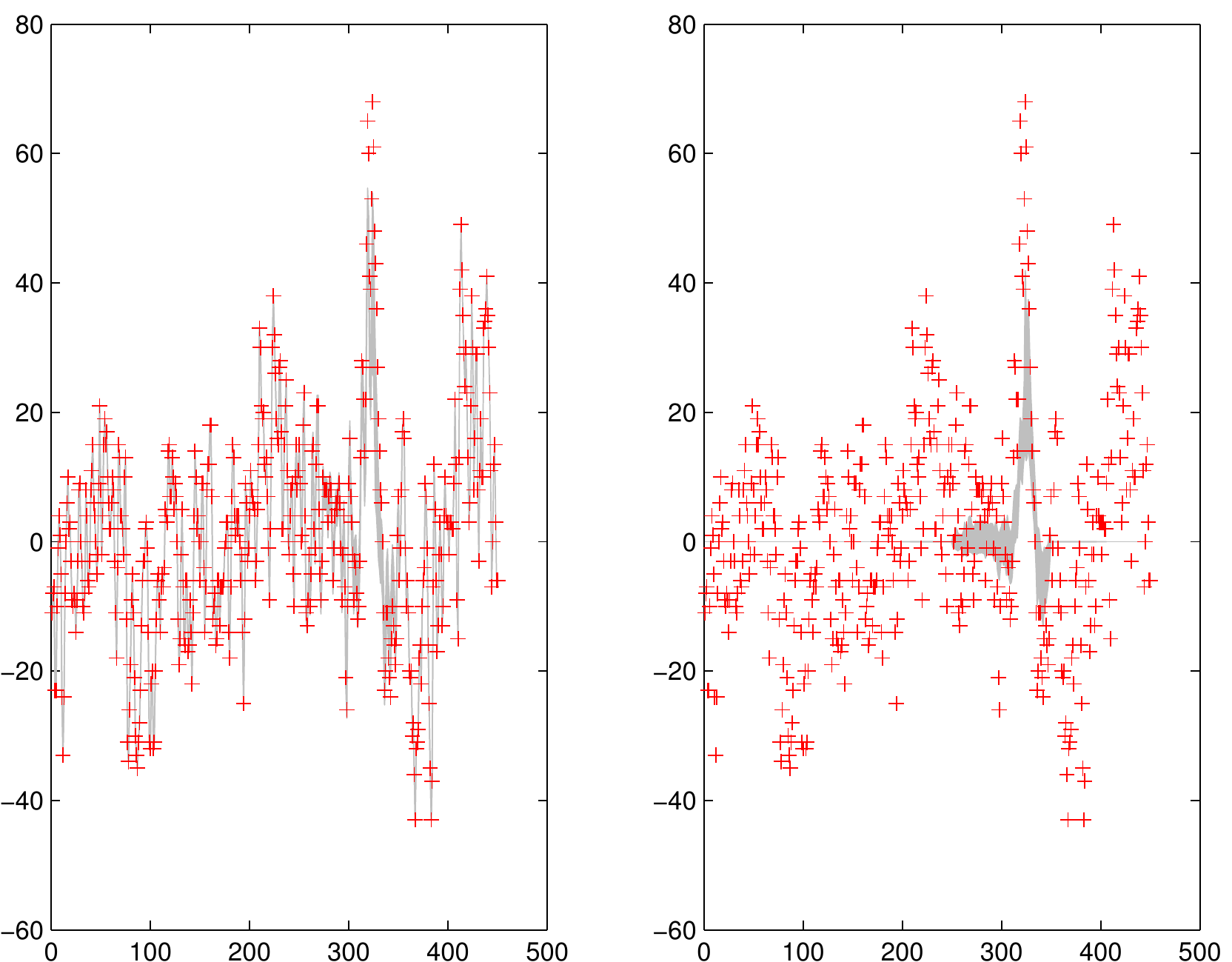}
\end{center}
\caption{\label{eeg_real} EEG signal (crosses) and 1 standard error
  confidence intervals for the EEG* (left panel) and EOG (right panel)
  signals obtained using the GP mixture model approach.}
\end{figure}

\section{Conclusions}
\label{sec:conclusions}
This paper has presented a novel approach for non-stationary Gaussian
processes across boundaries (i.e. changepoints for 1D signals).  It
builds piece-wise stationary prior covariance matrices from stationary
Gaussian process kernels and, where appropriate, asserts function
continuity or continuity of any function derivative at the region
boundaries.  The approach has been successfully demonstrated on sensor
fault detection and recovery and also on EEG signal tracking.

\bibliographystyle{plain}
\bibliography{reece_09_anomaly_removal}

\end{document}